\documentclass[11pt]{article}

\usepackage[in]{fullpage}

\usepackage{amsfonts,amscd,float}
\usepackage{array,booktabs,paralist}
\usepackage[rflt]{floatflt}
\usepackage{times,graphicx}
\usepackage{amsmath,amssymb,amsopn,algorithm,algorithmic,float,bbm,bm,enumerate,color,multirow}
\usepackage{setspace}
\usepackage{rotating}
\usepackage{array}	
\usepackage{hyperref}
\usepackage{url}
\usepackage{wrapfig}

\usepackage{subfig}
\usepackage[utf8]{inputenc} 
\usepackage[T1]{fontenc}    
\usepackage{hyperref}       
\usepackage{url}            
\usepackage{booktabs}       
\usepackage{amsfonts}       
\usepackage{nicefrac}       
\usepackage{microtype}      
\usepackage{amsmath}
\usepackage{amssymb}
\usepackage{bm}

\usepackage{caption}

\begin{document}

\title{R2N2: Residual Recurrent Neural Networks \\for Multivariate Time Series Forecasting}

%

\author{Hardik Goel\\
Dept of Computer Science \& Engineering\\
University of Minnesota, Twin Cities\\
goelx033@umn.edu
\and
Igor Melnyk\\
IBM Research, T. J. Watson Research Center\\
Yorktown Heights, NY\\
igor.melnyk@ibm.com
\and
Arindam Banerjee\\
Dept of Computer Science \& Engineering\\
University of Minnesota, Twin Cities\\
banerjee@cs.umn.edu}

\maketitle

\vspace*{-5mm}
\begin{abstract}
Multivariate time-series modeling and forecasting is an important problem with numerous applications. Traditional approaches such as VAR (vector auto-regressive) models and more recent approaches such as RNNs (recurrent neural networks) are indispensable tools in modeling time-series data.
In many multivariate time series modeling problems, there is usually a significant linear dependency component, for which VARs are suitable, and a nonlinear component, for which RNNs are suitable. Modeling such times series with only VAR or only RNNs can lead to poor predictive performance or complex models with large training times.
%
%
In this work, we propose a hybrid model called R2N2 (Residual RNN), which first models the time series with a simple linear model (like VAR) and then models its residual errors using RNNs. R2N2s can be trained using existing algorithms for VARs and RNNs. Through an extensive empirical evaluation on two real world datasets (aviation and climate domains), we show that R2N2 is competitive, usually better than VAR or RNN, used alone. We also show that R2N2 is faster to train as compared to an RNN, while requiring less number of hidden units.
\end{abstract}

\vspace*{-4mm}
\section{Introduction}
\label{sec:introduction}
\vspace*{-4mm}
Multivariate time-series modeling and forecasting constitutes an important problem with numerous applications in several real-world domains such as healthcare, finance, climate, and aviation \cite{lipton2015learning,mmbo16,taylor2007modelling}. As a result, classical time-series models
including autoregressive models like Vector Autoregression (VAR) \cite{lutkepohl07} and Autoregressive Integrated Moving Average (ARIMA)\cite{lutkepohl07}, latent state based models like Kalman Filters (KF) \cite{kalman} and others have been extensively studied and  used for multivariate time-series analysis.
While these models have been helpful in several domains, they have certain key limitations: they assume a linear dependence over time, and they are not well suited to model long-term dependencies.
Another practical limitation while using models like VAR is the issue of scalability and order selection, where order denotes the assumed duration of historical dependency. When fitting a VAR model, an appropriate order selection grid-search has to be run to select the best possible order for the data. For a $p$ dimensional dataset spanning $T$ timesteps, a VAR model of order $k$ requires $O(Tk^{2}p^{4} + k^{3}p^{6})$ time to estimate its $O(kp^{2})$ parameters. Now if the model is operating in a  high-dimensional setting along with long term dependencies, grid search for order selection can become computationally expensive.

To model non-linear temporal data, recent years have seen considerable success based on Recurrent Neural Networks (RNNs) and its variants (Gated Recurrent Units (GRUs) \cite{chung2014empirical}, Long Short-Term Memory (LSTMs) \cite{Hochreiter}) \cite{bahdanau2014neural, NIPS2008_3449, sutskever2014sequence}. While most of this work has been for discrete sequential data, such models have been used for continuous data as well for time-series prediction (see Section \ref{sec:related}). RNNs are non-linear models with the capability of modeling long-term dependencies without the need to explicitly specify the exact lag/order. Further, the number of parameters of an RNN does not depend on the order that the data exhibits, rather is mostly dependent on the dimension of the data. Thus these models do not require an extensive grid-search procedure to find the best order. However, there are issues with using RNNs alone for time-series as well. First, they are difficult to train well and may suffer from local minima problems \cite{lipton2015critical}. Even after carefully tuning the backpropagation algorithm, one might not reach the most optimal parameters. The second issue is that if the data has a significant linear component, the added non-linear complexity in RNNs might actually produce worse results as compared to the linear models \cite{zhang2003time}. Another issue is the training time of RNNs, which can be significant for large and complex models.

Motivated by these issues, in this work we propose R2N2s (Residual RNNs). Most real-world time series data can be assumed to have a linear dependency component and a non-linear component which can also capture long term dependencies. To handle this, R2N2s first fit a simple linear model, like an order-1 VAR, to the data and then use the error residuals from the linear model as input data to an RNN. Thus, the RNN only focuses on modeling the remaining, non-linear part of the data in the VAR's residuals, along with any long-term dependencies. Having two stages  effectively lessens the burden on the RNN while improving accuracy and using less training time.

We draw real-world data from two domains: aviation industry (multiple aircraft sensor measurements over time, for millions of flights) and climate science (indices recording the El Nino Southern Oscillation over several decades), and perform extensive empirical evaluation on them. The results illustrate multiple advantages of R2N2s: (1) order selection does not have to be performed for VAR; R2N2 takes care of the long-term dependency in the data, (2) less number of hidden units are required in the RNN component of R2N2 to achieve the same or even better performance than using RNN alone, (3) the training time for R2N2 becomes significantly less as compared to training RNN alone, and (4) there is improvement in the prediction performance as compared to using VAR or RNN separately. As a model, the initial component of R2N2 need not be a VAR, and other base models such as Kalman Filter (KF), Autoregressive Conditional Heteroscedastic (ARCH) models and variants can also be used. Hence, R2N2 is applicable to different kinds of datasets, and can take advantage of the existing, simple, domain specific (linear) models as the first component in R2N2.

The rest of the paper is organized as follows. Section \ref{sec:related} presents a discussion of the existing literature in time-series prediction using VARs and RNNs. In section \ref{sec:models}, we discuss the component models VAR, RNNs and LSTMs. Following that, in section \ref{sec:r2n2}, we describe our model in detail and present experimental results in section \ref{sec:experiments}. Finally, we conclude the paper in section \ref{sec:conc}.

\section{Related Work}
\label{sec:related}

{\bf Vector Auto-Regressive Models (VARs):}
VAR models \cite{lutkepohl07} and their extensions are arguably among the most widely used family of statistical models for multivariate time series forecasting. These models have been effective in a wide variety of applications, ranging from describing the behavior of economic and financial time series \cite{tsay05, simseco} to modeling dynamical systems  \cite{ljung98} and estimating brain function connectivity \cite{valdes05}, among others. Recent work has successfully applied VAR models to analyze multivariate aviation time series data, i.e., multiple sensor measurements for flights \cite{meba16,mmvb16,mmbo16}. By modeling the  time evolution of multiple on-board sensors, these approaches aimed at accurate prediction of aircraft behavior to discover anomalous events from a large dataset of flight information.

{\bf Recurrent Neural Networks (RNNs):} Recurrent Neural Networks (RNN) have been effective in modeling time-series data. Particularly, Long Short-Term Memory (LSTM) networks, which are a special kind of RNNs, originally introduced in \cite{Hochreiter} have shown remarkable performance with sequential data. Because of their ability to learn long term patterns in sequential data, they have recently been applied to a diverse set of problems, including handwriting recognition \cite{NIPS2008_3449}, machine translation \cite{sutskever2014sequence}, text classification \cite{dai2015semi}, in medical diagnosis \cite{lipton2015learning} and many others. Most of the existing work, however, focuses on discrete sequential data and/or classification tasks. For continuous-valued time series modeling, the literature is somewhat limited, and includes the work of \cite{wollmer2008abandoning} for emotion recognition and \cite{weninger2014line} for online music mood regression. Apart from this, the work of \cite{Schmidhuber2005} has used LSTM-based prediction model on the Mackey Glass time-series, achieving promising results. Some of the more recent work in \cite{2017arXiv170404110F} has focused explicitly on multivariate time-series using RNNs to perform probabilistic time-series prediction.

{\bf Hybrid Models:}
Since our proposed model (R2N2) is a type of hybrid ensemble model, i.e., models which combine two or more different types of base models in suitable ways, we review some of the existing work on hybrid ensembles in this section. In \cite{zhang2003time}, a residual model using Autoregressive Integrated Moving Average (ARIMA) and a feedforward neural network is presented. The idea is that real world time-series data is a mixture of linear and non-linear parts. ARIMA models the linear part while the neural network models its residuals. Another form of this hybrid modeling is explored in \cite{pai2005hybrid}, using ARIMA and Support Vector Machines (SVM) on univariate time-series data. However, both these works ignore the time dependency in ARIMA's residuals and treat them as independent data samples. In contrast, R2N2 takes into account the time dependency of the residuals through the use of RNNs. Yet another work \cite{ardalani2010chaotic}, on hybrid residual modeling uses a combination of Elman and NARX neural networks. Here an Elman network is used to model the original data, followed by multiple levels of residual modeling using more Elman networks and finally a NARX network to combine all the previous outputs. Though the time dependence in the residuals is not ignored, but unlike R2N2, they do not consider a combination of a linear/short term and nonlinear/long term dependencies. Another recent work \cite{lai2017modeling}, proposes a new kind of model that has a recurrent, a convolutional and an autoregressive (AR) component, all trained together using gradient descent to perform time series prediction. Though this work combines an AR and an RNN component like us, the difference in our work is that the AR component is a not a part of the core model. Rather we look at modeling the residual of the AR part using RNNs, which leads to a different direction of work. In spirit, R2N2 is trained sequentially like gradient boosting, although we consider a hybrid ensemble and the model is for multivariate time series forecasting. 


\section{Existing Models: VAR, RNN, LSTM}
\label{sec:models}
{\bf Vector Auto-Regressive Models:}
Vector AutoRegression (VAR) is a popular statistical approach to model linear dependencies among multiple features that evolve in time \cite{ljung98,lutkepohl07}. In general form, the $k$-th order VAR model can be written as
\vspace*{-2mm}
\begin{align}
\label{eq:var}
x_t = &A_1x_{t-1} + \ldots + A_kx_{t-k} + \epsilon_t,
\end{align}
\noindent where $A\in\mathbb{R}^{p\times p}$ are the matrices of coefficients, $x\in\mathbb{R}^{p}$ is the vector of parameters, $\epsilon\in\mathbb{R}^{p}$ is zero-mean white noise,  and $t=k+1, \ldots, T$, where $T$ denotes the length of time series.  The subscript $k$ determines the lag of the model, i.e., the degree to which the data in the current time step depends on the data in the past. 
To estimate the VAR parameters, the model in \eqref{eq:var} is usually transformed into a form suitable for a least-square estimator. 

{\bf Recurrent Neural Networks:}
\begin{figure*}
\captionsetup[subfigure]{font=small,aboveskip=2pt,belowskip=-10pt}
\subfloat[\footnotesize{Recurrent Neural Network}]{\label{suppfig:rnn}\includegraphics[width=0.5\linewidth]{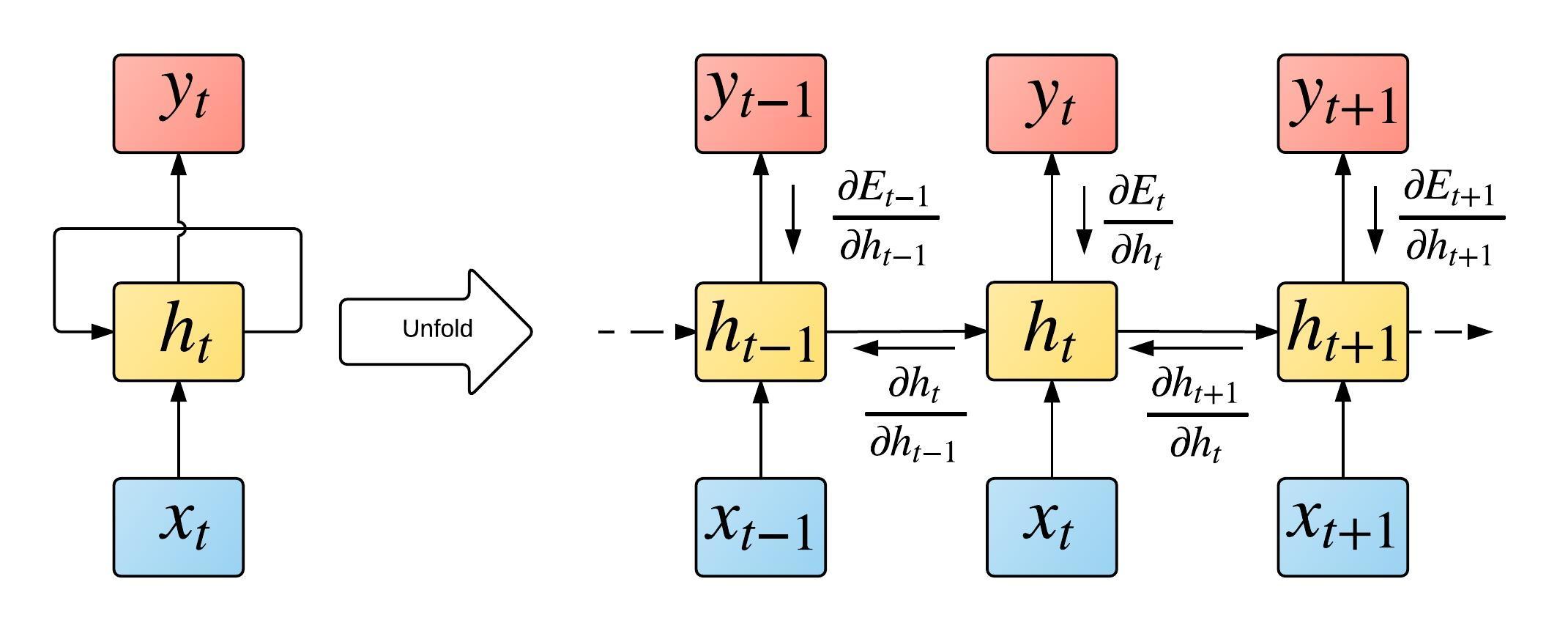}}
\subfloat[\footnotesize{An LSTM cell at time $t$}]{\label{suppfig:lstm}\includegraphics[width=0.5\linewidth]{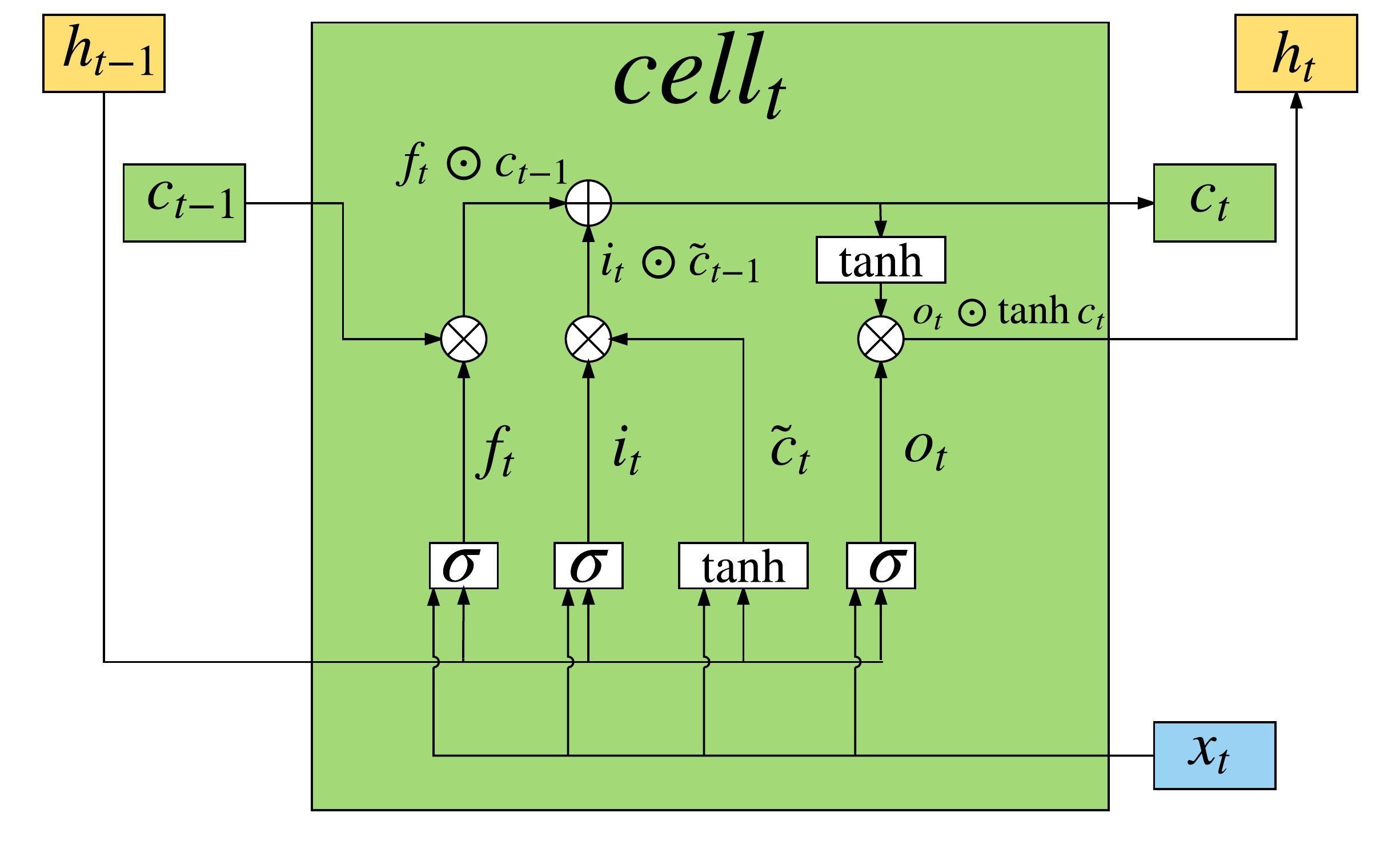}}



\vspace{2mm}
\caption{(a)The left side of the figure illustrates the feedback loop of the model, where $x_{t}$, $h_{t}$ and $y_{t}$ are the input, hidden state and output. The right side shows RNN unfolded in time, which also illustrates the flow of the gradients during the backpropagation step of the training; $E_{t}$ is the error in $y_{t}$ at time $t$. (b) The color mapping is similar to (a), showing that $h_{t}$ here is equivalent to hidden state $h_{t}$ of RNN and $c_{t}$ is a new component that is added to the LSTM cells. The $h_{t}$ here can be used as the cell output or converted using a feed-forward connection to an output $y_{t}$. The $\sigma$s represent the gates of the cell, which are single layer neural networks with sigmoidal activation.}
\label{suppfig:rnndesc}
\end{figure*}
Given a time-series data $x_{1}, x_{2},\dots, x_{t}$, a Recurrent Neural Network (RNN) is defined by the following recurrent relation
\vspace*{-2mm}
\begin{align}
\label{suppeq:rnn}
h_{t} = \sigma(Wx_{t} + Uh_{t - 1} + b),
\end{align}
where, $x_{t} \in \mathbb{R}^{p}$ is the input at time $t$, $W \in \mathbb{R}^{n \times p}$, $U \in \mathbb{R}^{n \times n}$, $b \in \mathbb{R}^{n}$ are the hidden state parameters, $h_{t}$ and $h_{t - 1} \in \mathbb{R}^{n}$ are the hidden state vectors at times $t$ and $t - 1$, respectively, and $\sigma$ is the logistic sigmoid function. Figure \ref{suppfig:rnn} shows an example of RNN model, whose main part is the feedback loop, generating the hidden state $h_{t}$ at time step $t$ from the current input $x_t$ and previous hidden state $h_{t - 1}$.

The main issue with RNN model is that during training, while applying the backpropagation-through-time (BPTT) algorithm, the error gradients can quickly vanish/explode \cite{pascanu2013difficulty}. This is due to the application of the chain rule of the differentiation at each time step. This prevents the network from capturing the long-term dependencies in the data. LSTMs, an extension of RNNs, were introduced as a mechanism to prevent these issues by backpropagating a constant error gradient.

{\bf Long Short Term Memory:}
To address the vanishing gradient problem of RNN, the LSTM defines a new hidden state, called a cell. Each cell has its own \emph{cell state}, which acts like a memory, and various control mechanisms, called gates, enable modification of the cell memory. Through the use of such mechanisms, the LSTM can effectively learn when to forget the old memories (\emph{forget gate}), when to add new memories from the current input (\emph{input gate}) and what memories to present as output from the current cell (\emph{output gate}).

Each gate is a single layer neural network whose weights are the extra parameters to be learned during training. The gates have sigmoidal activation in their outputs, squashing the output to $[0, 1]$ range. These values indicate, as fractions, how much reading, writing or forgetting to perform, giving increased learning power to the model.

An LSTM-based RNN is governed by the following set of equations,
\begin{eqnarray}
f_{t} &=& \sigma(W_{f}x_{t} + U_{f}h_{t - 1} + b_{f}) \\
\label{suppeq:lstm.it}
i_{t} &=& \sigma(W_{i}x_{t} + U_{i}h_{t - 1} + b_{i}) 
\label{suppeq:lstm.ctilde} \\
\tilde{c}_{t} &=& \tanh(W_{c}x_{t} + U_{c}h_{t - 1} + b_{c}) \\
\label{suppeq:lstm.ct}
c_{t} &=& i_{t} \odot \tilde{c}_{t} + f_{t} \odot c_{t - 1} \\
\label{suppeq:lstm.ot}
o_{t} &=& \sigma(W_{o}x_{t} + U_{o}h_{t - 1} + b_{o}) \\
\label{suppeq:lstm.ht}
h_{t} &=& o_{t} \odot \tanh(c_{t})
\end{eqnarray}
where, $f_{t}, i_{t}, \tilde{c}_{t}, c_{t}, o_{t}$ and $h_{t} \in \mathbb{R}^{n}$, represent the outputs of forget gate, input gate, candidate state, cell state, output gate and the final cell output, respectively. The $W$s, $U$s and $b$s are the parameters of the network that are learned during training. $\odot$ represents the element-wise Hadamard product. Note that $h_{t}$ itself can be used as the final output of the network. If $h_{t}$ does not have the desired dimension, one can use a fully connected feed-forward neural network to convert it to an output $y_{t}$ with the desired dimensions. Figure \ref{suppfig:lstm} illustrates the entire flow inside the LSTM cell.

The main reason LSTM does not suffer from the vanishing gradient problem is because \emph{cell state} $c_{t}$ is updated using only addition operation, as opposed to an RNN, which involves a sigmoidal transformation \eqref{suppeq:rnn}. Thus during backpropagation-through-time, only a constant error is propagated back to every step. Due to these reasons, LSTM is able to learn long-range dependencies in the data.

\section{Residual Recurrent Neural Networks}
\label{sec:r2n2}
\begin{figure*}[t]
\vspace*{-1mm}
\captionsetup[subfigure]{font=small,aboveskip=0pt,belowskip=-5pt}
\subfloat[\footnotesize{A general R2N2}]{\label{fig:r2n2_block}\includegraphics[width=.4\textwidth]{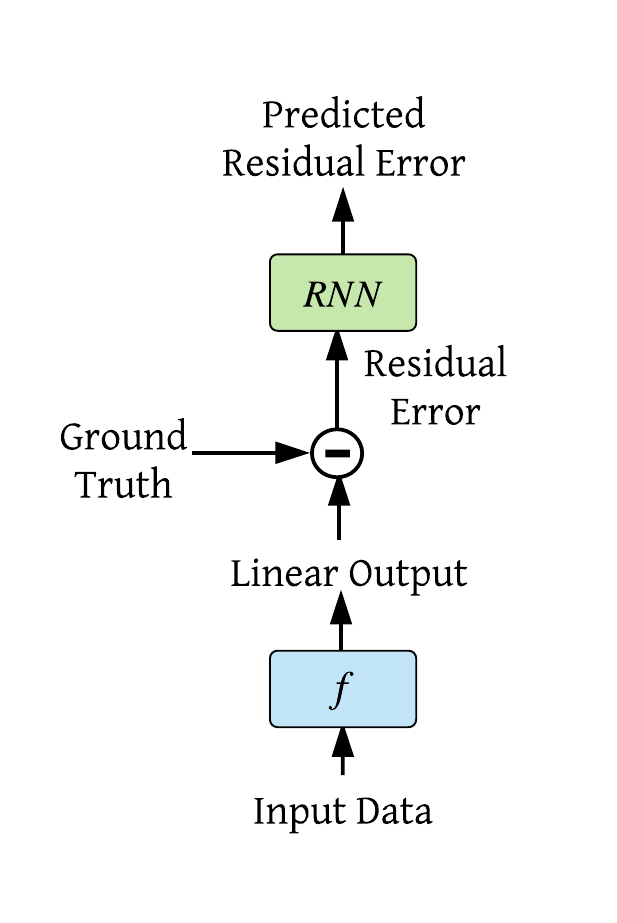}}
\subfloat[\footnotesize{A specific R2N2 realization}]{\label{fig:r2n2_lstm}\includegraphics[width=.57\textwidth]{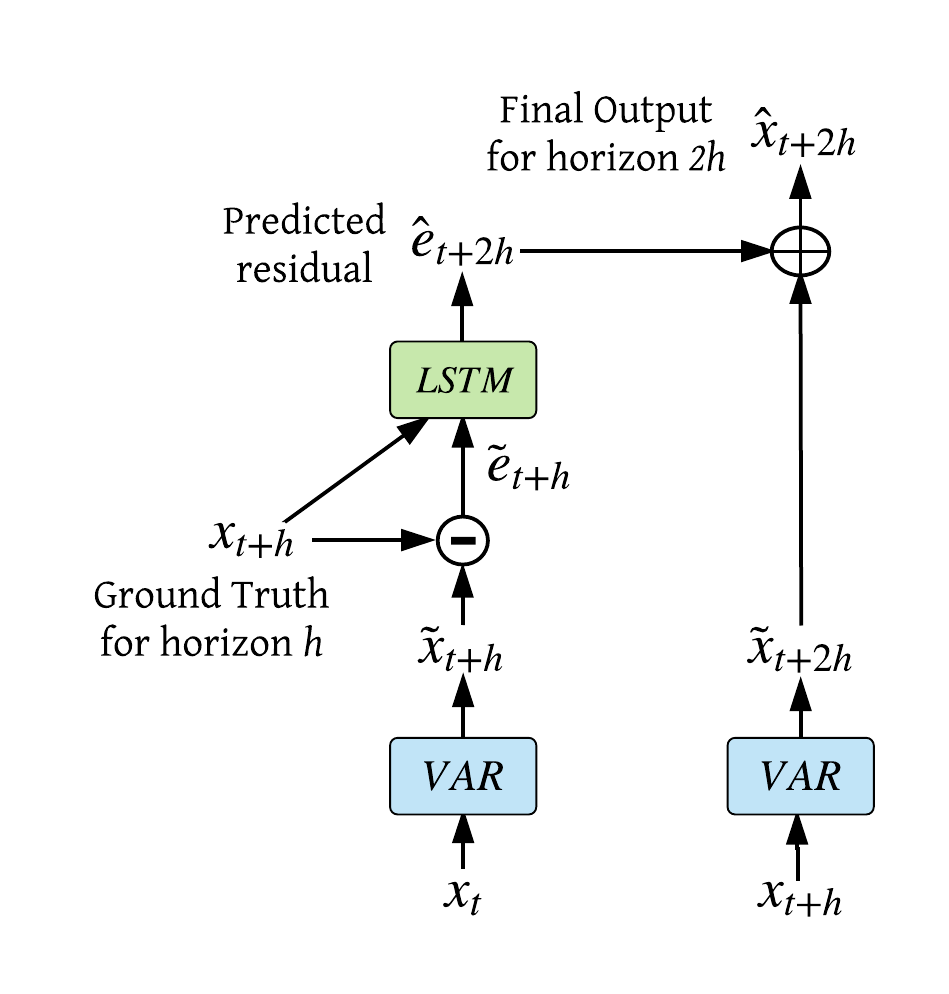}}
%
\caption{(a) An illustration of a general R2N2 model. The block $f$ represents the base model, which is a linear time-series model like VAR or Kalman Filter. The predictions from this model are compared with the ground truth to generate residuals. These residuals are then fed into the RNN model as input. The RNN is then tasked with predicting the error that the base model $f$ will make in the future. This predicted residual can then be added to $f$'s predictions, to generate the final corrected outputs. (b) One possible realization of R2N2 with VAR as linear model and an LSTM as the non-linear model. This is the model that we have used in our experiments.}
\label{fig:r2n2}
\vspace*{-4mm}
\end{figure*}


In this section, we describe our proposed Residual Recurrent Neural Network (R2N2). Several real-world multivariate time-series data usually consist of some aspects which can be captured by a linear model and certain other aspects which need a nonlinear possibly long memory model. Classical models for time series analysis, such as VAR and variants \cite{kalman, lutkepohl07} do well for linear models which have short memory, i.e., where a small order VAR suffices. However, such models were not designed to capture nonlinear and potentially long-memory dependencies. RNNs are suitable for such nonlinear and potentially long-memory dependencies. However, RNNs usually need large training sets and long training times. 

The idea behind R2N2 is simple. Given that several real-world multivariate time-series models do have both linear and nonlinear dependencies, R2N2 uses a linear model to capture the linear dependencies, and the residual errors of the linear model are used to train an RNN. Thus, R2N2 has both a linear and a nonlinear component with the linear part tasked with modeling the linear dependencies and the RNN focusing on the residual errors of the linear model, which are nonlinear by design. While one can train an RNN directly on the original time series, a large part of the modeling effort may get devoted to getting the linear part right, which is usually a significant component of several real-world time series data. R2N2 lets the component RNN focus only on the residual error of the linear model, which may hopefully lead to faster training and simpler models, e.g., smaller number of hidden units. As we show in Section~\ref{sec:experiments}, R2N2 does show these advantages quite consistently over a full RNN trained on the original time series. 


Figure \ref{fig:r2n2_block} shows a high level illustration of the R2N2 model for multivariate time-series forecasting for some specific horizon $h$. The horizon is the amount of time in the future the prediction is being made for. This can vary based on the prediction's usefulness for the specific application. In the figure, $f$ represents a base (linear) model, such as VAR or variants, which takes in the original input data and makes the first level of predictions. These predictions are compared against the ground truth at that point to generate the residuals. The multi-variate residual time-series would contain nonlinearities and long-term dependencies that the (linear) base model could not effectively model. The nonlinear model, which is a Recurrent Neural Network (RNN), takes over at this point. The RNN is tasked with learning to predict the error that the base model will make while predicting the next horizon's output. To do this, the RNN receives as input the entire error time series from the base model until the current time step and possibly also the original input data. Once the RNN has predicted the error for the next horizon, this can be combined with $f$'s prediction, to generate the model's final prediction. Note that for inputs where the linear model is adequate, the residual error will be small, and the RNNs have little to do; on the other hand, if the linear model yields large errors, the RNN effectively takes over the modeling.

This general idea can be easily instantiated according to the requirements of the domain and computational power (see Figure~\ref{fig:r2n2_lstm}). For instance, if the time series is $p$-dimensional and believed to have linear dependence on previous steps in time, the base linear model can be a collection of $p$ Autoregressive (AR) models or a single Vector AR (VAR) model. On the other hand, if there seems to be a latent state governing the data, a Kalman Filter (KF) can be substituted into the base model. For the non-linear RNN component, variants of RNNs such as Gated Recurrent Unit (GRU) or Long Short-Term Memory (LSTM) models can be used as they have demonstrated a good long term pattern learning capacity. Also, the residual error time series from the base model's predictions can be further augmented using the original input time series to enable the RNN to make better predictions. This can be useful because the RNN gets to know what the true data looked like for a particular error made by the base model.
In terms of equations, the R2N2 can be described as below
\vspace*{-1mm}
\begin{eqnarray}
\label{eq:r2n2.1}
\tilde{x}_{t+h} &=& f(x_{:t})~, \\
\label{eq:r2n2.2}
\tilde{e}_{t+h} &=& x_{t+h} - \tilde{x}_{t+h}~, \\
\label{eq:r2n2.3}
\hat{e}_{t+2h} &=& RNN([\tilde{e}_{:t+h}, x_{:t+h}])~, \\
\label{eq:r2n2.4}
\hat{x}_{t+2h} &=& f(x_{:t+h}) + \hat{e}_{t+2h}~,
\end{eqnarray}

\vspace*{-2mm}
where we use $:\!\!t$ or $:\!\!t+h$ to denote that data up to that time step can be used.
Here \eqref{eq:r2n2.1} represents the underlying linear base model denoted by $f$. The linear model takes input $x_{:t}$ up to time step $t$ and generates the output $\tilde{x}_{t+h}$, which is the model's prediction for a horizon of $h$. In \eqref{eq:r2n2.2}, $x_{t+h}$ refers to the ground truth value at time $t + h$, using which we get the linear model's residual $\tilde{e}_{t+h}$. An RNN is used as the secondary component to model the residuals. Equation \eqref{eq:r2n2.3} shows that the RNN takes as input the residual $\tilde{e}_{:t+h}$ up to time $t+h$ augmented with the original time-series $x_{:t+h}$. Using these inputs, the RNN is tasked with predicting how much error the linear model will make in its next prediction for time step $t+2h$. This prediction ($\hat{e}_{t+2h}$) is then combined with the linear model's prediction $f(x_{:t+h})$ to produce the final model output as shown in \eqref{eq:r2n2.4}. The subtraction and addition in the above equations are both elementwise operations on vectored quantities.

We present a specific realization of the R2N2 model in Figure \ref{fig:r2n2_lstm} with VAR as the base model and an LSTM as the non-linear model. This is the realization that we have used in our experiments for this work. No changes are required to the existing algorithms for training such a model, i.e., the VAR can be trained using traditional least square estimators \cite{lutkepohl07}, while the LSTMs are trained using gradient descent \cite{Hochreiter}. This enables us to use the existing body of work on training such models while providing a novel combination of the two.

\section{Experiments}
\label{sec:experiments}
In this section, we describe our datasets, experimental setup and discuss the results using the following models: \textbf{VAR}, \textbf{RNN} (vanilla LSTM) and \textbf{R2N2} (Residual RNN with LSTM).

\subsection{Datasets}
\label{datasets}
\noindent {\bf Aviation Data.} The first dataset is the Flight Operations Quality Assurance (FOQA) dataset from NASA \cite{nasadata}. This data contains air traffic flight sensor information and is used to detect issues in aircraft operation due to mechanical, environmental or human factors \cite{mmbo16,statler2003nasa}. The data contains over a million flights, each having a record of about 700 (multivariate) time series measurements, sampled at 1 Hz over the duration of the flight. These parameters include both discrete and continuous readings from control switches (like thrust, autopilot, flight director, etc.) and sensors (like altitude, angle of attack, drift angle, etc.). For our experiments, we selected 42 features representing all the continuous sensor measurements from 160 flights of the same type of aircraft, landing at the same airport. The focus was on a portion of the flight below $10,000$ feet until touchdown (duration $600$-$1500$ timestamps), which makes for about 120,000 time-steps of data in all. We split it into training, validation and 5 test sets with 100, 10 and 10 flights in each test set, respectively.

\noindent {\bf ENSO Data.} Our second dataset represents the El Nino Southern Oscillation (ENSO) data. The ENSO cycle refers to the coherent and sometimes very strong year-to-year variations in sea-surface temperatures, rainfall, surface air pressure and atmospheric circulation that occur across the equatorial Pacific Ocean \cite{enso}. The dataset consists of 7 indices such as NINO 1+2, NINO 3.4, etc., with monthly measurements from 1950 to 2008. Each feature represents sea surface temperature measurements (or a combination of those) across different regions. In all, this consists of 707 multivariate measurements, which are split into training set (60\%), validation set (20\%) and test set (20\%).

\subsection{Experimental Details}
The aviation data was normalized by z-scoring so that it has zero mean and unit variance. We perform 1-step ahead prediction for all the variables, which means the prediction horizon $h = 1$. All the hyperparameter selection was done using a held-out validation set and the results reported are the mean performance on the 5 held-out test sets.

In the case of the ENSO dataset, we perform a 6 month ahead prediction, which means that the minimum possible lag of VAR was 6. We denote this as VAR-1 for this dataset. Since this is a climate dataset, there is an annual periodic structure observed in some of the features. To account for this, we subtract the monthly mean of the training data from the entire time series, before training the models, which helps in producing better predictions. The results are reported after converting the predictions back to the original scale. Also, since this is a very small dataset, we needed to heavily regularize the RNNs with $L2$ norm of the recurrent weights to prevent overfitting. The hyperparameter selection was again done on the validation set and the results are reported for the test set.

For both datasets, the best VAR was found by increasing the lag until the error stopped decreasing, while at the same time grid-searching for the best regularization parameter from \{0.05, 0.5, 5.0, 50.0, 500.0\}. R2N2 always uses VAR-1 as the base model. For the RNNs and R2N2s, we used the LSTM-based cell and determined the best size of hidden layer by grid-search. As for the depth of the RNNs, we found that a single layer performed the best and was the fastest to train as opposed to making the network deeper, which did not result in any more gains. The Adam \cite{kingma2014adam} algorithm was used to train the RNNs. The learning rate was reduced by a factor of 10 whenever the validation loss stopped improving. This resulted in better performance while also preventing overfitting. The training was stopped when the learning rate went below a certain threshold and no further improvements were observed in the validation loss.

We used two evaluation metrics: Mean Relative Squared Error (MRSE) defined as
\begin{equation}
MRSE = \frac{\sqrt[]{\sum_{i=1}^{n} \sum_{t=1}^{T}(X_{it} - \hat{X}_{it})^{2}}}{\sqrt[]{\sum_{i=1}^{n} \sum_{t=1}^{T}(X_{it} - mean(X_i))^{2}}}  \label{eq:eval.mrse}~,
\end{equation}
and and  Relative Error (RE) defined as
\begin{equation}
RE = \frac{\sqrt[]{\sum_{i=1}^{n} \sum_{t=1}^{T}(X_{it} - \hat{X}_{it})^{2}}}{\sqrt[]{\sum_{i=1}^{n} \sum_{t=1}^{T}(X_{it})^{2}}} \label{eq:eval.re}~,
\end{equation}
where $\bm{X}, \bm{\hat{X}} \in \mathbb{R}^{n \times T}$ are the ground truth and the model predictions respectively. $X_{it}$ represents the value of the $i$th dimension at time $t$. Lower values are better for both the metrics. MRSE measures the goodness of the model's prediction as compared to predicting the feature's mean all the time. RE measures the error relative to the original data's scale.

\begin{figure*}[t]
\vspace*{-3mm}
\captionsetup[subfigure]{font=small,aboveskip=2pt,belowskip=-10pt}
\subfloat[\footnotesize{Prediction sample for an aviation feature}]{\label{fig:flightdata.predictions.sfig1}\includegraphics[width=.5\textwidth]{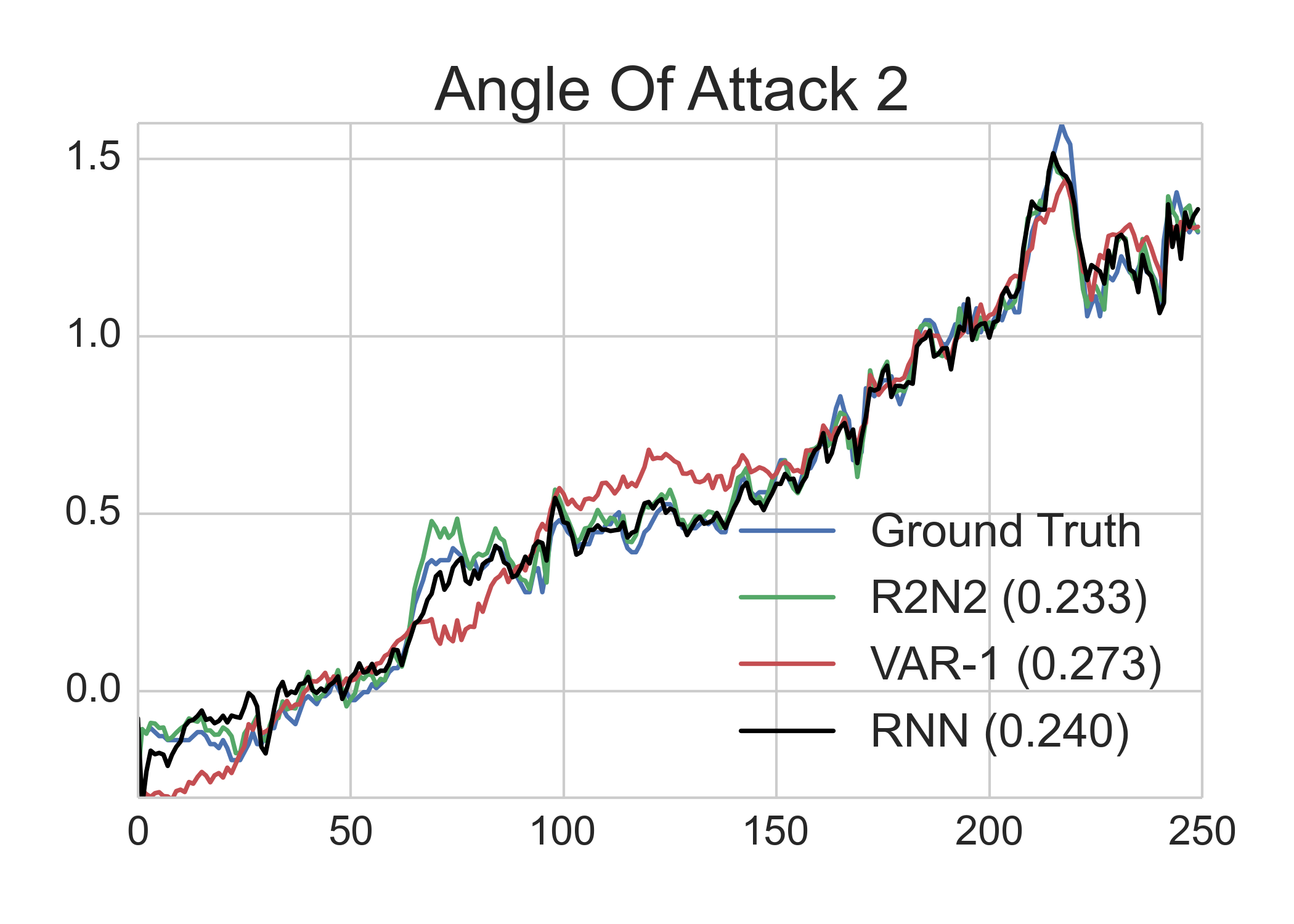}}
\subfloat[\footnotesize{Prediction sample for another aviation feature}]{\label{fig:flightdata.predictions.sfig4}\includegraphics[width=.5\textwidth]{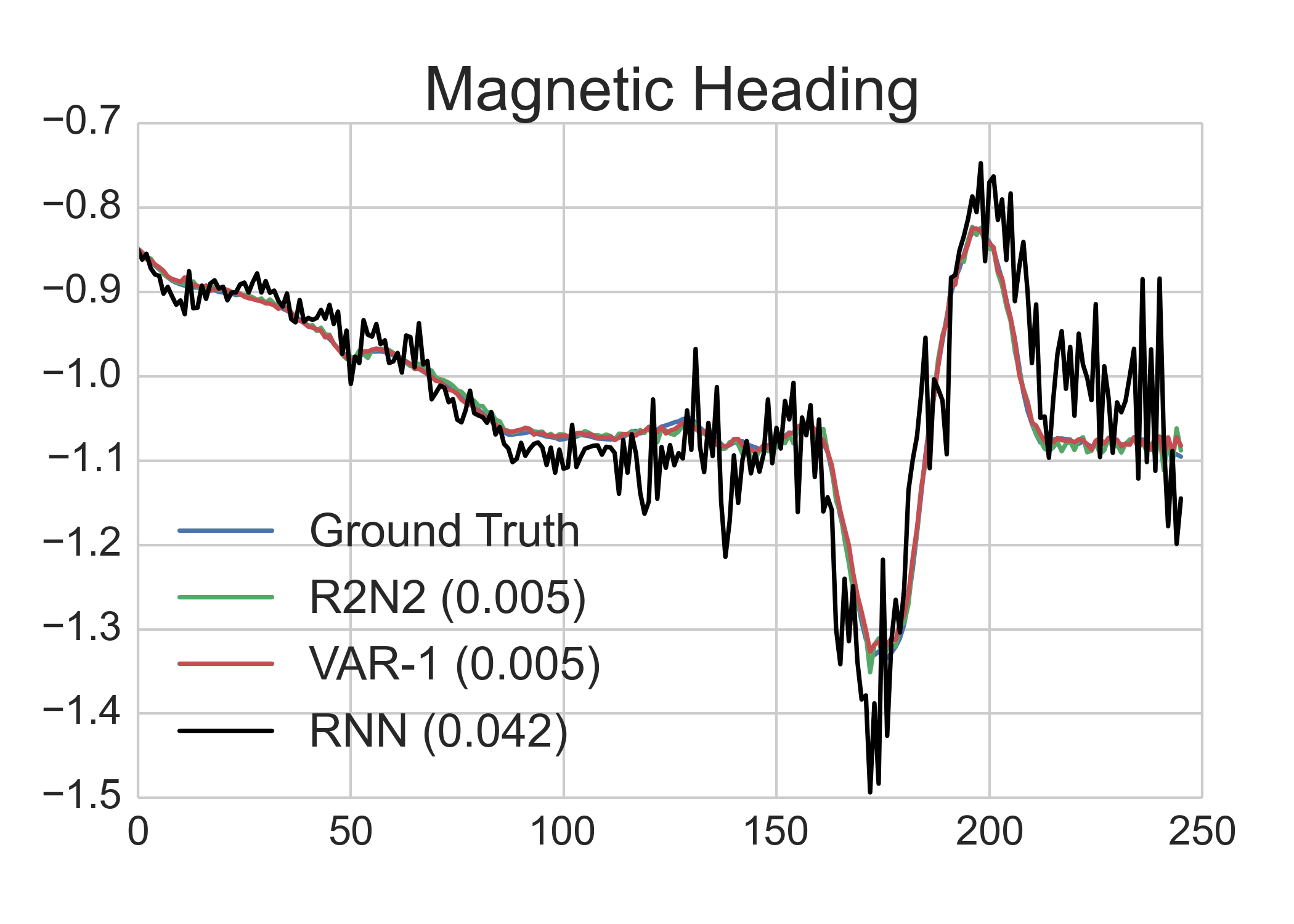}}

\caption{Sample prediction plots of selected features from the aviation dataset. The values in parentheses in the plot legends are the MRSE for that feature (same is true for all other plots). Note that for the left plot, which is a feature with high volatility, VAR does poorly but R2N2 consistently improves its errors. Also, R2N2 and RNN are comparable in performance. For the right plot, which is relatively ``smooth'' or less volatile feature, VAR performs rather well. R2N2 still tries to improve a bit, while RNN's prediction is very volatile.}
\label{fig:aviation.pred}
\vspace*{-3mm}
\end{figure*}

\subsection{Results}
\subsubsection{Qualitative predictions}
\vspace*{-1mm}
The prediction plots for selected features from a randomly selected flight for the aviation dataset are shown in Figure \ref{fig:aviation.pred} (equivalent plots for all the 42 features are included in the supplementary material). The figure shows the ground truth and the predictions of VAR-1, RNN and R2N2 (with VAR-1 as the base model). R2N2 is doing well for both highly volatile and ``smooth'' features, while RNN struggles with smooth features and VAR-1 with volatile features. This was noticed for other features as well. Thus in some sense, VAR-1 can be thought of as providing stability to R2N2, while still giving it enough flexibility to model high volatility.

\begin{figure*}
\vspace*{-13mm}
\subfloat[\footnotesize{Predictions for an ENSO index}]{\label{fig:flightdata.predictions.sfig1}\includegraphics[width=.5\textwidth]{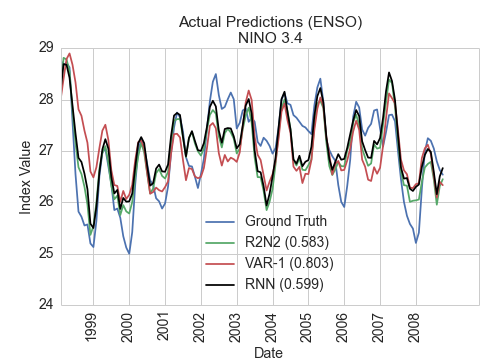}}
\subfloat[\footnotesize{Predictions for another ENSO feature}]{\label{fig:flightdata.predictions.sfig4}\includegraphics[width=.5\textwidth]{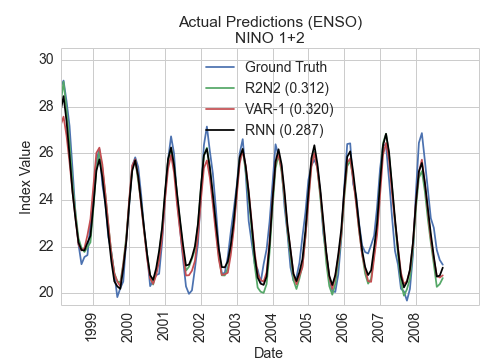}}
\\
\captionsetup[subfigure]{font=small,aboveskip=2pt,belowskip=-10pt}
\subfloat[\footnotesize{Prediction residuals for an ENSO index}]{\label{fig:flightdata.predictions.sfig1}\includegraphics[width=.5\textwidth]{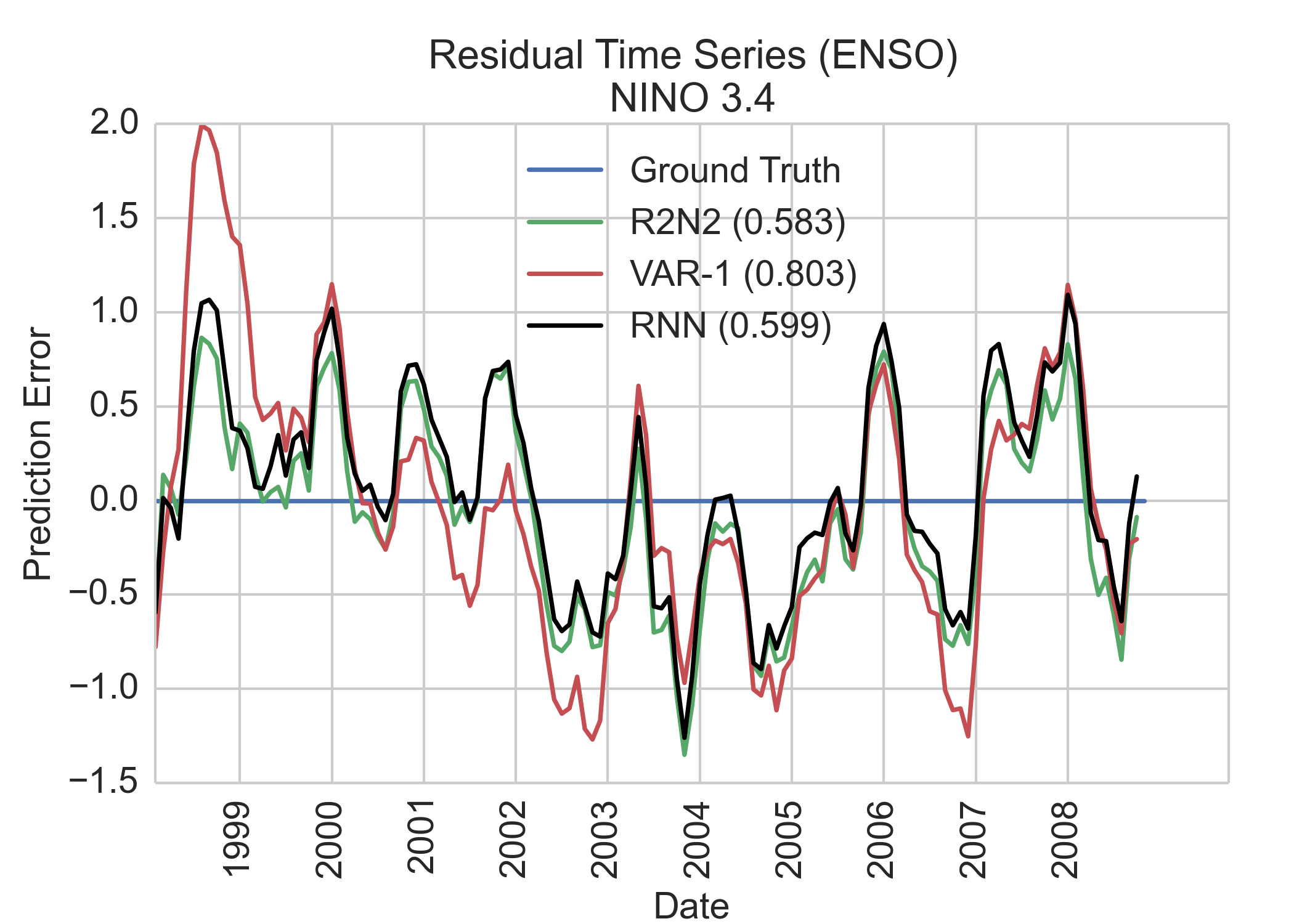}}
\subfloat[\footnotesize{Prediction residuals for another ENSO index}]{\label{fig:flightdata.predictions.sfig4}\includegraphics[width=.5\textwidth]{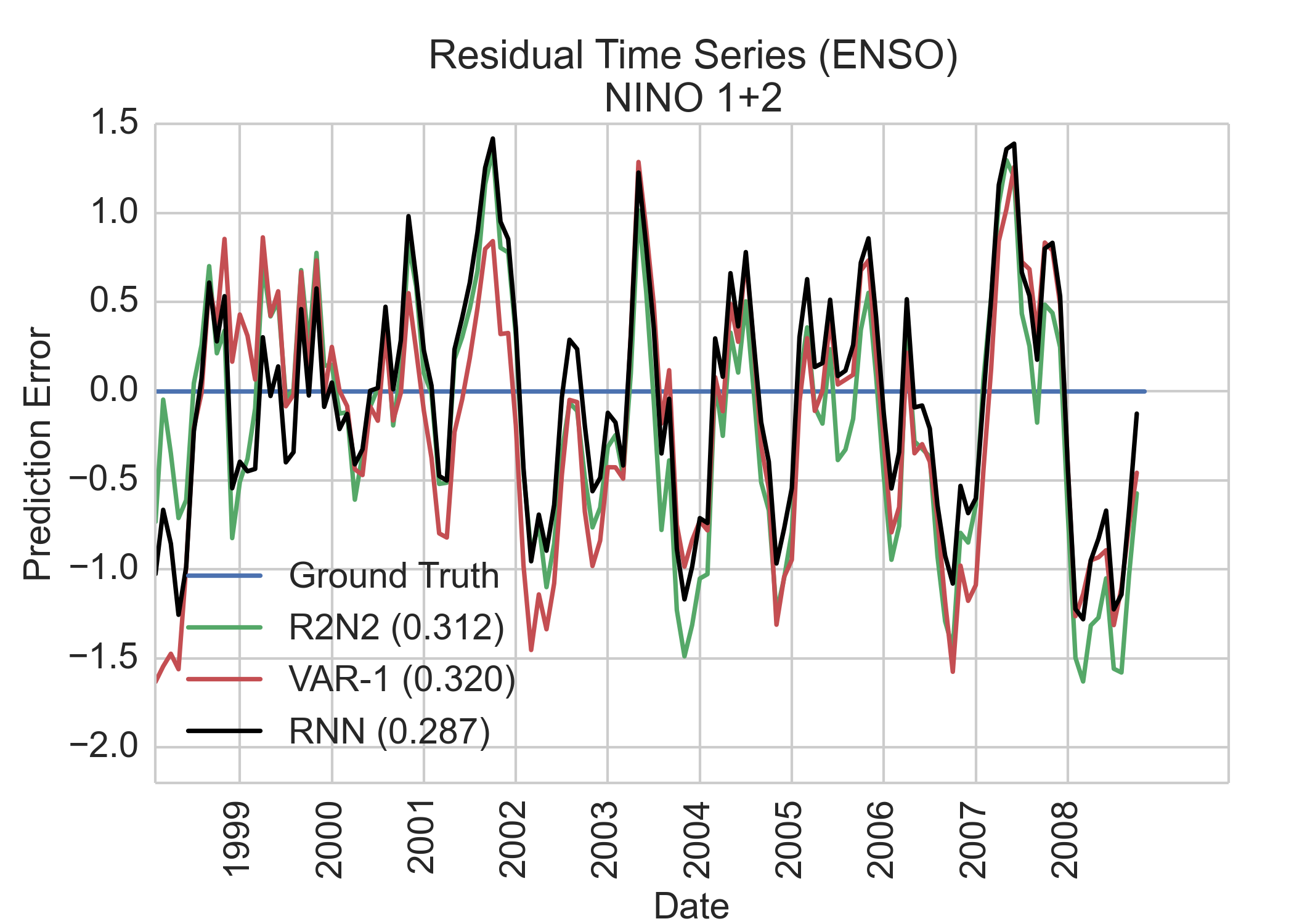}}


\caption{Sample prediction plots of selected features from the ENSO dataset. The top row contains the actual data and model predictions. R2N2 does best for the first index (NINO 3.4), while RNN does the best in the case of the second (NINO 1+2), which is highly periodic. The bottom row shows the plots of the residuals for each model's predictions as compared to the ground truth (which is 0 for no error). Note that R2N2 reduces the errors made by VAR-1. Also, for NINO 1+2, the plot shows that RNN wins but R2N2 is close to RNN and away from VAR. }
\vspace*{-3mm}
\label{fig:enso.pred}
\end{figure*}

Figure \ref{fig:enso.pred} shows the actual prediction and prediction residual plots for selected indices from the ENSO dataset (equivalent plots for all the 7 indices are included in the supplementary material). On the 7 indices from the ENSO dataset, RNN and R2N2 both significantly outperform VAR, which is expected because these indices are highly non-linear phenomena. For 2 indices out of 7, RNN outperforms R2N2. These indices (NINO 1+2 and NINO 3), show highly periodic structure because they are affected by seasons. After subtracting the monthly mean, the RNN is able to focus well on modeling the remaining part (which in itself is a very simple residual model). On the other hand, for 4 of the indices, which do not have such a periodic structure, the mean subtraction does not help much and R2N2 continues to do well. Figure \ref{fig:enso.pred} also shows the plots of the prediction residuals in the bottom row. Note that, for NINO 1+2, though RNN does the best, R2N2 is still moving away from VAR in its predictions and closer to RNN. This suggests that R2N2 is, in a sense, good at capturing the ``best of both worlds''. Overall, among the three methods, R2N2 is either the best or the second best across both the datasets and all features.

\subsubsection{Aggregate quantitative performance}
\begin{figure}[t]
\centering
\vspace*{-3mm}
\captionsetup[subfigure]{font=small,aboveskip=2pt,belowskip=-10pt}
\subfloat[\footnotesize{MRSE for aviation data}]{\label{fig:aviation.mrse}\includegraphics[width=.33\textwidth]{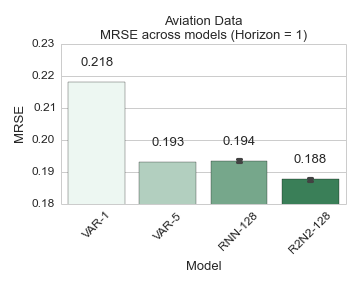}}
\subfloat[\footnotesize{MRSE for ENSO data}]{\label{fig:enso.mrse}\includegraphics[width=.33\textwidth]{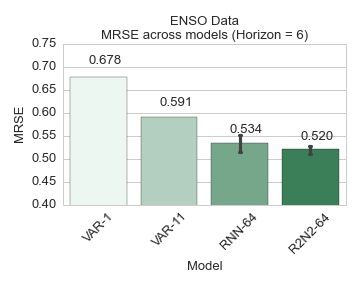}}
\subfloat[\footnotesize{MRSE across hidden units}]{\label{fig:aviation.rse.hidden}\includegraphics[width=.33\textwidth]{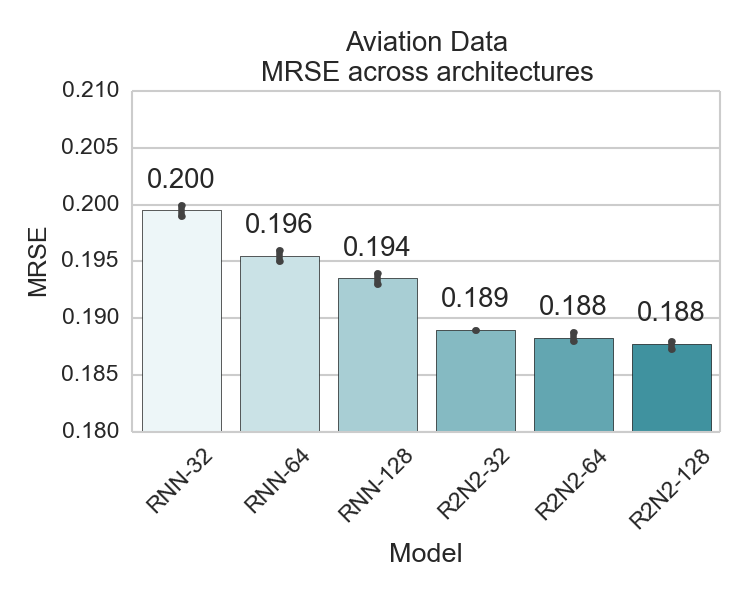}}


\caption{(a) Mean MRSE results on aviation test sets for all 3 models - VAR, RNN and R2N2. Note that for VAR we show both order-1 results and the best order results (which was 5 in this case). RNN-128 means an RNN with a hidden layer size of 128. R2N2-128 uses VAR-1 as the base model and an RNN with hidden layer size of 128. R2N2-128 performs the best among all the model architectures.  The error bars in both the figures represent the variability due to different initializations of parameters across multiple runs. The bars are small indicating good stability in performance for both RNN and R2N2. (b) MRSE results on the ENSO data for all the models. R2N2-64 performs the best among all the model architectures. (c) MRSE values across different hidden layer sizes of RNN and R2N2 for the aviation data. Note that even R2N2-32 performs better than RNN-128. This implies that very low complexity RNNs can be used as components in R2N2, which provides savings in terms of computational complexity.}
\label{fig:aviation.rse}
\end{figure}

\vspace*{-1mm}
Figure \ref{fig:aviation.mrse} shows the mean values of MRSE that we achieved on the aviation test sets. R2N2-128 (i.e. R2N2 with VAR-1 as base model and RNN with hidden layer size 128) performs the best. VAR-5 performs at par with RNN-128, which indicates that proper order selection with regularization is necessary for VARs to achieve good performance. Also, note that VAR-1's MRSE is high as compared to other models, but we still propose to use it as the base model for R2N2. If a higher order VAR (like VAR-5) is used instead as a part of R2N2, no improvements are noticed over the performance of VAR-5. A possible explanation might be that a well-fitted VAR, when used as the base model, biases R2N2 towards a ``VAR operating domain'' and the R2N2 is not able to make further improvements. Using the simpler VAR-1 provides a good base model for R2N2 while at the same time giving enough flexibility to the RNN part. This also frees up the R2N2 model from having to perform an order selection for the VAR component. Figure \ref{fig:enso.mrse} shows the same results for the ENSO dataset after converting the predictions back to the original scale by adding monthly means of indices. The RE results follow similar trend as MRSE, so in the interest of space, they are shown in the supplementary material.

\subsubsection{Profiling training times : RNN vs R2N2}
\begin{figure}[t]
\centering
\subfloat[\footnotesize{Learning curve for $H = 32$}]{
\label{fig:timing_32}\includegraphics[width=.33\textwidth]{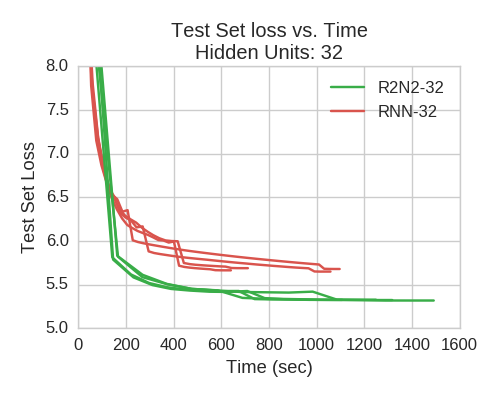}
}
\subfloat[\footnotesize{Learning curve for $H = 64$}]{
\label{fig:timing_64}\includegraphics[width=.33\textwidth]{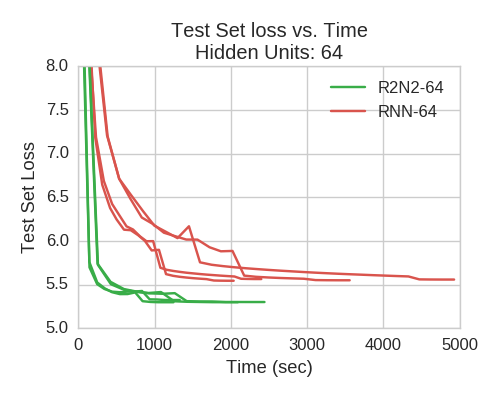}
}
\subfloat[\footnotesize{Learning curve for $H = 128$}]{
\label{fig:timing_128}\includegraphics[width=.33\textwidth]{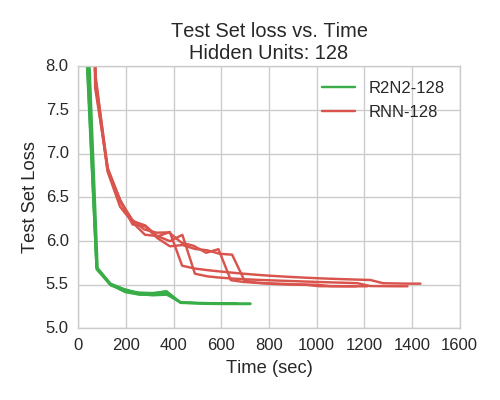}
}
\vspace*{1mm}
\caption{These plots depict the progression of the loss on the aviation test sets as the training iterations proceed for 4 different initializations of every model. Note that R2N2s train much faster than RNNs. Also, there is a steep drop in R2N2's loss in the beginning due to the assistance of VAR.}
\label{fig:aviation.timing.results}
\end{figure}

\vspace*{-1mm}
Next we study the time it takes to train RNNs and R2N2s across different architectures on the aviation dataset. By training time for R2N2 in this section, we refer to the training time of the RNN component of R2N2, while training time of RNN refers to that of the RNN-only model. The training time of the VAR portion of R2N2 is negligible (orders of magnitude less) as compared to the RNNs, so we skip mentioning it separately in this discussion. Figure \ref{fig:aviation.timing.results} shows learning curves for different sizes of the hidden layer for RNN and R2N2. The figure makes it clear that R2N2s are very fast in achieving the minimum test set error as opposed to plain RNNs. For R2N2s, we can observe a steep drop in the beginning, which can be attributed to the base VAR model. Right after the first iteration, R2N2 figures out that a lot its work has already been done for it. In contrast, for RNNs we can see that the loss gradually decreases over a number of iterations before it flattens out at a final test set loss higher than R2N2.
This experiment verifies what we suggested earlier that using the VAR as the base model in R2N2 lessens the burden on the RNN component. It is comparatively much easier for the RNN component to quickly model the VAR's residuals and converge faster.

\subsubsection{Architecture comparison : RNN vs R2N2}
\vspace*{-1mm}
In this experiment, we study the effect of modifying the architecture of the RNNs on the prediction performance, using the aviation dataset. Again, when we talk about RNN it refers to the RNN-only model, while R2N2 refers to the RNN component of the R2N2. Figure \ref{fig:aviation.rse.hidden} shows the effect of modifying the number of nodes in the hidden layer on the MRSE. The performance consistently improves as the hidden layer size increases for both RNN and R2N2. However, a more interesting thing to note is that even a 32-node hidden layer in R2N2 performs better than a 128-node hidden layer in the RNN. This result shows that because of using VAR as the base linear model, the input to the RNN becomes much less complex and can be modeled with rather simpler architectures. Combined with the results of the previous subsection on timing profile, this allows for two-fold reduction (less complex model and faster convergence) in training time while providing better performance. The RE results follow similar trend as MRSE, due to lack of space, they are in the supplementary material.


\section{Conclusions}
\label{sec:conc}
In this work, we presented a new model, R2N2, for multivariate time-series prediction by combining the traditional time-series models with Recurrent Neural Networks. The idea is to use the existing machinery of classical time-series analysis to model the time-series and then use the RNN to model the residuals. We explore a specific realization of this model with VAR as the base linear model and an LSTM-based RNN as the non-linear component. Through extensive experimentation on two real world datasets from the aviation and climate domains, we show that R2N2s achieve better prediction performance than both VAR and RNN individually. Further we also show through experiments that R2N2s require simpler neural network architectures and hence provide significant reduction in training times. Another advantage of R2N2 is that it can be easily customized to different domains by using domain-specific base linear models like Kalman Filters, ARCH models, etc., paving the direction for future work.


{\bf Acknowledgements:} This work was supported by NSF grants IIS-1563950, IIS-1447566, IIS-1447574, IIS-1422557,
CCF-1451986, CNS-1314560, IIS-0953274, IIS-1029711, and NASA grant NNX12AQ39A.


\bibliographystyle{plain}


\begin{thebibliography}{10}

\bibitem{enso}
{El-Nino Southern Oscillation}. {A}vailable at
  https://www.esrl.noaa.gov/psd/data/climateindices/.

\bibitem{nasadata}
{NASA Flight Dataset}. {A}vailable at
  {https://c3.nasa.gov/dashlink/projects/85/}.

\bibitem{ardalani2010chaotic}
Muhammad Ardalani-Farsa and Saeed Zolfaghari.
\newblock Chaotic time series prediction with residual analysis method using
  hybrid elman--narx neural networks.
\newblock {\em Neurocomputing}, 73(13):2540--2553, 2010.

\bibitem{bahdanau2014neural}
Dzmitry Bahdanau, Kyunghyun Cho, and Yoshua Bengio.
\newblock Neural machine translation by jointly learning to align and
  translate.
\newblock {\em arXiv preprint arXiv:1409.0473}, 2014.

\bibitem{chung2014empirical}
Junyoung Chung, Caglar Gulcehre, KyungHyun Cho, and Yoshua Bengio.
\newblock Empirical evaluation of gated recurrent neural networks on sequence
  modeling.
\newblock {\em arXiv preprint arXiv:1412.3555}, 2014.

\bibitem{dai2015semi}
A.~M. Dai and Q.~V. Le.
\newblock Semi-supervised sequence learning.
\newblock In {\em NIPS}, pages 3079--3087, 2015.

\bibitem{2017arXiv170404110F}
V.~{Flunkert}, D.~{Salinas}, and J.~{Gasthaus}.
\newblock {DeepAR: Probabilistic Forecasting with Autoregressive Recurrent
  Networks}.
\newblock {\em ArXiv e-prints}, April 2017.

\bibitem{NIPS2008_3449}
A.~Graves and J.~Schmidhuber.
\newblock Offline handwriting recognition with multidimensional recurrent
  neural networks.
\newblock In {\em NIPS}, pages 545--552. 2009.

\bibitem{Hochreiter}
S.~Hochreiter and J.~Schmidhuber.
\newblock Long short-term memory.
\newblock {\em Neural Comput.}, 9(8):1735--1780, November 1997.

\bibitem{kalman}
Rudolph~Emil Kalman.
\newblock A new approach to linear filtering and prediction problems.
\newblock {\em Transactions of the ASME--Journal of Basic Engineering},
  82(Series D):35--45, 1960.

\bibitem{kingma2014adam}
Diederik Kingma and Jimmy Ba.
\newblock Adam: A method for stochastic optimization.
\newblock {\em arXiv preprint arXiv:1412.6980}, 2014.

\bibitem{lai2017modeling}
Guokun Lai, Wei-Cheng Chang, Yiming Yang, and Hanxiao Liu.
\newblock Modeling long-and short-term temporal patterns with deep neural
  networks.
\newblock {\em arXiv preprint arXiv:1703.07015}, 2017.

\bibitem{lipton2015learning}
Z.~C. Lipton, D.~C. Kale, C.~Elkan, and R.~Wetzell.
\newblock Learning to diagnose with lstm recurrent neural networks.
\newblock In {\em ICLR}, 2015.

\bibitem{lipton2015critical}
Zachary~C Lipton, John Berkowitz, and Charles Elkan.
\newblock A critical review of recurrent neural networks for sequence learning.
\newblock {\em arXiv preprint arXiv:1506.00019}, 2015.

\bibitem{ljung98}
L.~Ljung.
\newblock {\em System identification: theory for the user}.
\newblock Springer, 1998.

\bibitem{lutkepohl07}
H.~Lutkepohl.
\newblock {\em New introduction to multiple time series analysis}.
\newblock Springer, 2007.

\bibitem{meba16}
I.~Melnyk and A.~Banerjee.
\newblock Estimating structured vector autoregressive models.
\newblock In {\em ICML}, 2016.

\bibitem{mmbo16}
I.~Melnyk, B.~Matthews, A.~Banerjee, and N.~Oza.
\newblock Semi-{M}arkov switching vector autoregressive model-based anomaly
  detection in aviation systems.
\newblock {\em Knowledge Discovery and Data Mining}, 2016.

\bibitem{mmvb16}
I.~Melnyk, B.~Matthews, H.~Valizadegan, A.~Banerjee, and N.~Oza.
\newblock Vector autoregressive model-based anomaly detection in aviation
  systems.
\newblock {\em Journal of Aerospace Information Systems}, 13(4):161--173, 05
  2016.

\bibitem{pai2005hybrid}
Ping-Feng Pai and Chih-Sheng Lin.
\newblock A hybrid arima and support vector machines model in stock price
  forecasting.
\newblock {\em Omega}, 33(6):497--505, 2005.

\bibitem{pascanu2013difficulty}
R.~Pascanu, T.~Mikolov, and Y.~Bengio.
\newblock On the difficulty of training recurrent neural networks.
\newblock {\em ICML}, 28:1310--1318, 2013.

\bibitem{Schmidhuber2005}
J.~Schmidhuber, D.~Wierstra, and F.~Gomez.
\newblock {Evolino: Hybrid neuroevolution / optimal linear search for sequence
  learning}.
\newblock {\em IJCAI}, pages 853--858, 2005.

\bibitem{simseco}
Christopher~A. Sims.
\newblock Macroeconomics and reality.
\newblock {\em Econometrica}, 48(1):1--48, 1980.

\bibitem{statler2003nasa}
I.~C. Statler and D.~A. Maluf.
\newblock Nasa's aviation system monitoring and modeling project.
\newblock Technical report, SAE Technical Paper, 2003.

\bibitem{sutskever2014sequence}
I.~Sutskever, O.~Vinyals, and Q.~V. Le.
\newblock Sequence to sequence learning with neural networks.
\newblock In {\em NIPS}, pages 3104--3112, 2014.

\bibitem{taylor2007modelling}
S.~J. Taylor.
\newblock {\em Modelling financial time series}.
\newblock World Scientific Publishing, 2007.

\bibitem{tsay05}
R.~S. Tsay.
\newblock {\em Analysis of financial time series}, volume 543.
\newblock John Wiley \& Sons, 2005.

\bibitem{valdes05}
P.~A Vald{\'e}s-Sosa, J.~M S{\'a}nchez-Bornot, A.~Lage-Castellanos,
  M.~Vega-Hern{\'a}ndez, et~al.
\newblock Estimating brain functional connectivity with sparse multivariate
  autoregression.
\newblock {\em Philosophical Transactions of the Royal Society},
  360(1457):969--981, 2005.

\bibitem{weninger2014line}
F.~Weninger, F.~Eyben, and B.~Schuller.
\newblock On-line continuous-time music mood regression with deep recurrent
  neural networks.
\newblock In {\em IEEE ICASSP}, pages 5412--5416, 2014.

\bibitem{wollmer2008abandoning}
M.~W{\"o}llmer, F.~Eyben, et~al.
\newblock Abandoning emotion classes-towards continuous emotion recognition
  with modelling of long-range dependencies.
\newblock In {\em INTERSPEECH}, volume 2008, pages 597--600, 2008.

\bibitem{zhang2003time}
G~Peter Zhang.
\newblock Time series forecasting using a hybrid arima and neural network
  model.
\newblock {\em Neurocomputing}, 50:159--175, 2003.

\end{thebibliography}









\newpage
\appendix

\section{Qualitative prediction plots of aviation data}
\begin{figure*}
\vspace*{-12mm}
\centering
\captionsetup[subfigure]{font=small,aboveskip=0pt,belowskip=0pt,labelformat=empty}
\subfloat[]{\includegraphics[width=0.45\linewidth]{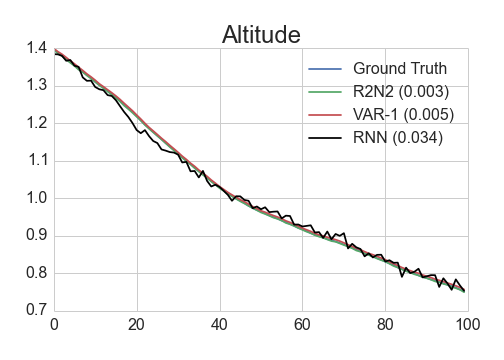}}
\subfloat[]{\includegraphics[width=0.45\linewidth]{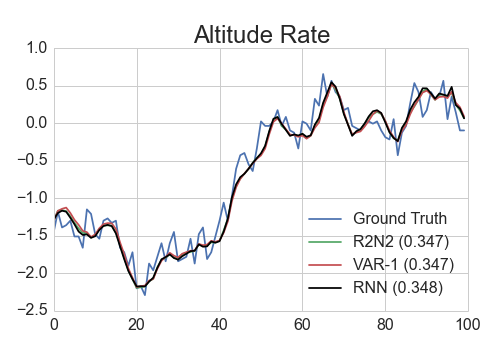}}
\vspace*{-12mm}
\\
\subfloat[]{\includegraphics[width=0.45\linewidth]{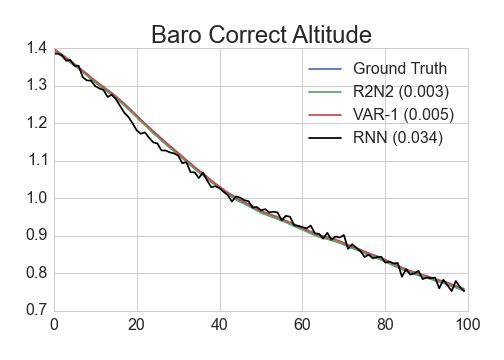}}
\subfloat[]{\includegraphics[width=0.45\linewidth]{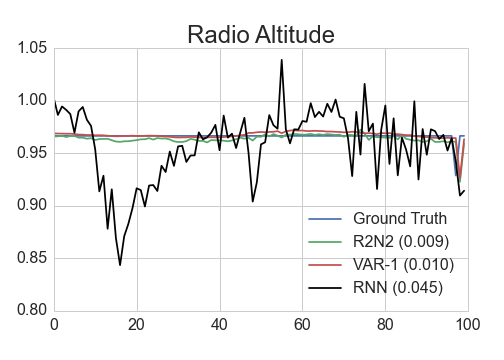}}
\vspace*{-12mm}
\\
\subfloat[]{\includegraphics[width=0.45\linewidth]{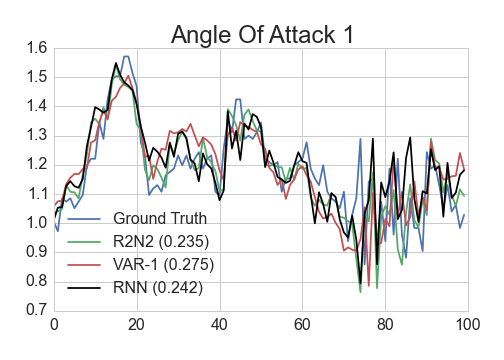}}
\subfloat[]{\includegraphics[width=0.45\linewidth]{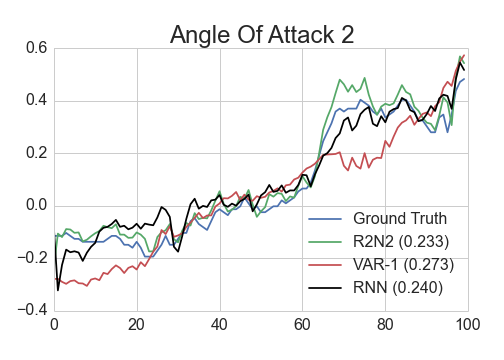}}
\vspace*{-12mm}
\\
\subfloat[]{\includegraphics[width=0.45\linewidth]{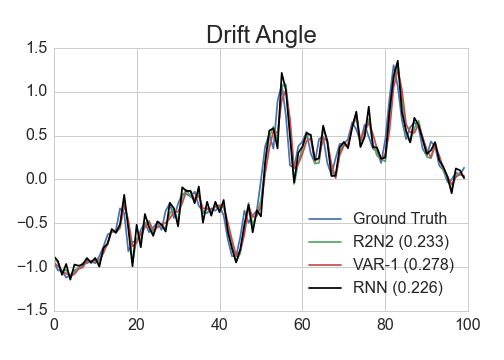}}
\subfloat[]{\includegraphics[width=0.45\linewidth]{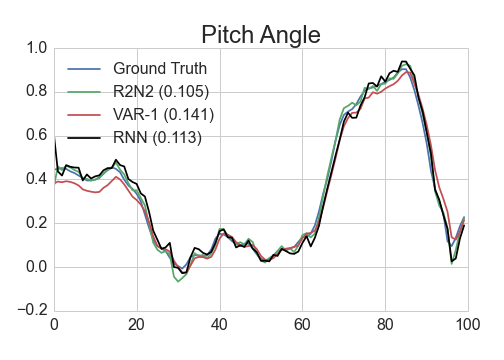}}
\vspace*{-8mm}
\caption{Plots of portions of features from a flight in the {\bf aviation data} and the 1-step predictions made by all models. Note that the x-axis and y-axis are different for each feature because selected sections of each feature have been zoomed-in for clarity. 8 features are shown here out of the total 42.}
\label{suppfig:aviation.pred.1}
\end{figure*}

\begin{figure*}
\vspace*{-12mm}
\centering
\captionsetup[subfigure]{font=small,aboveskip=0pt,belowskip=0pt,labelformat=empty}
\subfloat[]{\includegraphics[width=0.45\linewidth]{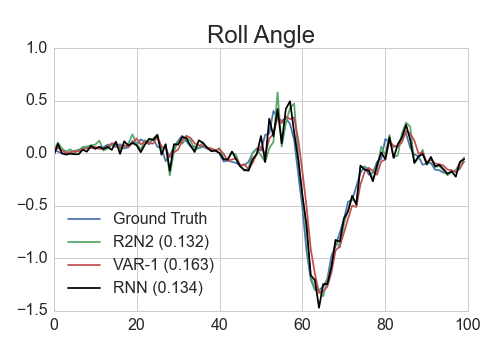}}
\subfloat[]{\includegraphics[width=0.45\linewidth]{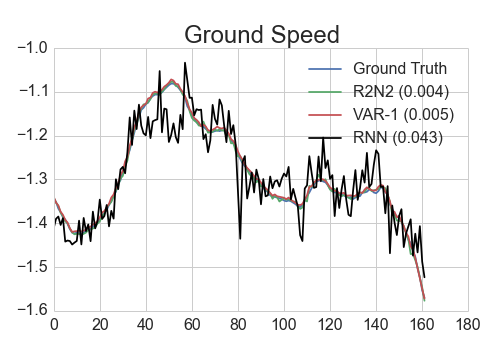}}
\vspace*{-12mm}
\\
\subfloat[]{\includegraphics[width=0.45\linewidth]{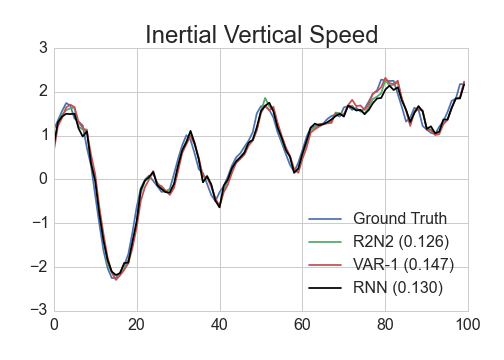}}
\subfloat[]{\includegraphics[width=0.45\linewidth]{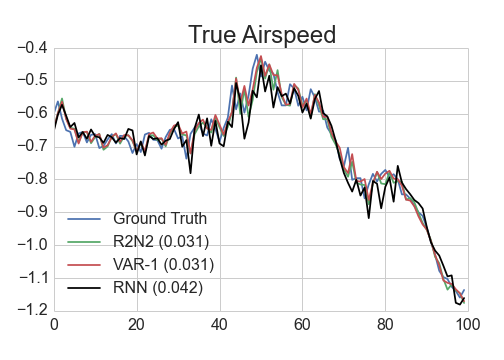}}
\vspace*{-12mm}
\\
\subfloat[]{\includegraphics[width=0.45\linewidth]{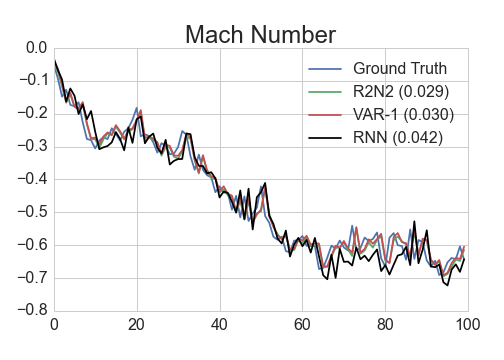}}
\subfloat[]{\includegraphics[width=0.45\linewidth]{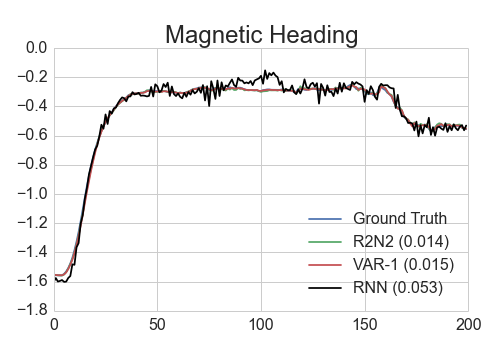}}
\vspace*{-12mm}
\\
\subfloat[]{\includegraphics[width=0.45\linewidth]{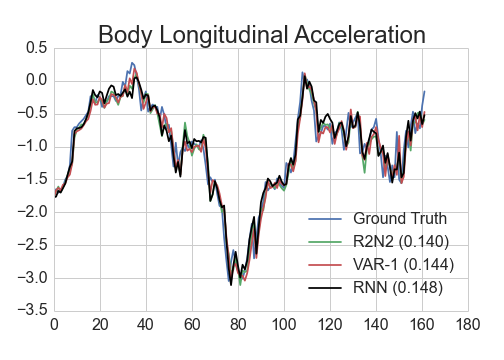}}
\subfloat[]{\includegraphics[width=0.45\linewidth]{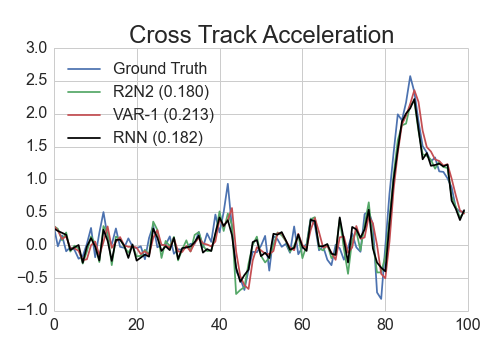}}
\vspace*{-8mm}
\caption{Plots of portions of features from a flight in the {\bf aviation data} and the 1-step predictions made by all models. Note that the x-axis and y-axis are different for each feature because selected sections of each feature have been zoomed-in for clarity. 8 features are shown here out of the total 42.}
\label{suppfig:aviation.pred.2}
\end{figure*}

\begin{figure*}
\vspace*{-12mm}
\centering
\captionsetup[subfigure]{font=small,aboveskip=0pt,belowskip=0pt,labelformat=empty}
\subfloat[]{\includegraphics[width=0.45\linewidth]{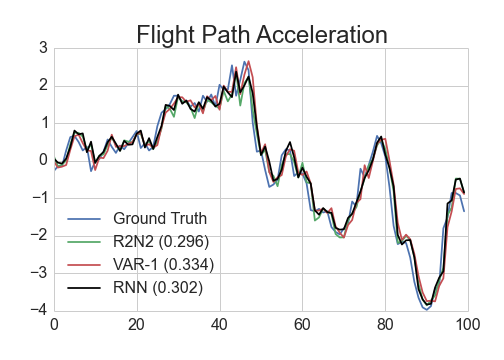}}
\subfloat[]{\includegraphics[width=0.45\linewidth]{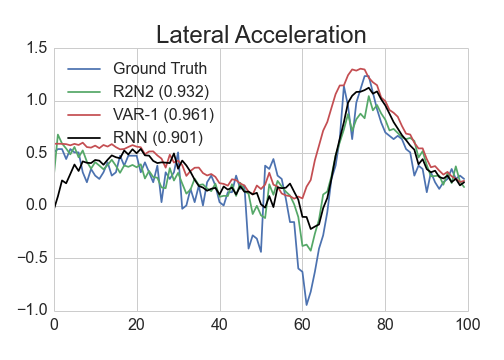}}
\vspace*{-12mm}
\\
\subfloat[]{\includegraphics[width=0.45\linewidth]{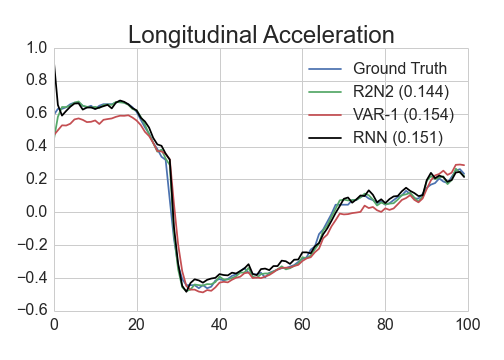}}
\subfloat[]{\includegraphics[width=0.45\linewidth]{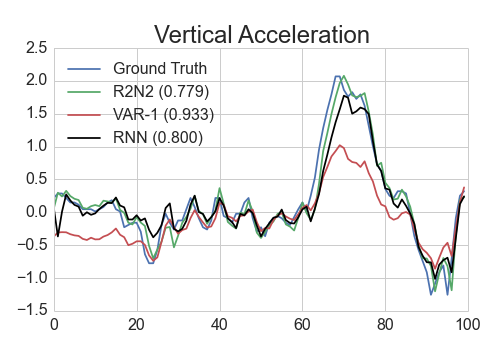}}
\vspace*{-12mm}
\\
\subfloat[]{\includegraphics[width=0.45\linewidth]{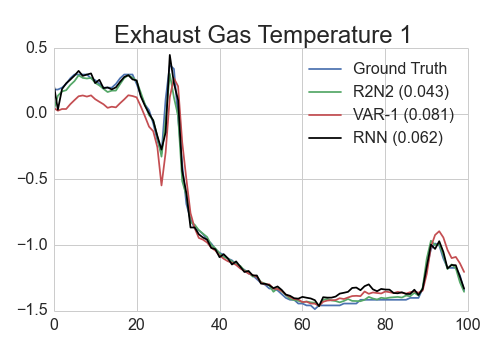}}
\subfloat[]{\includegraphics[width=0.45\linewidth]{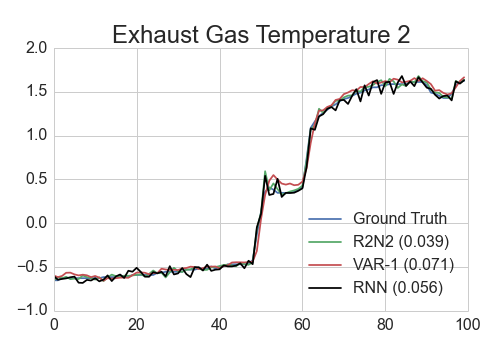}}
\vspace*{-12mm}
\\
\subfloat[]{\includegraphics[width=0.45\linewidth]{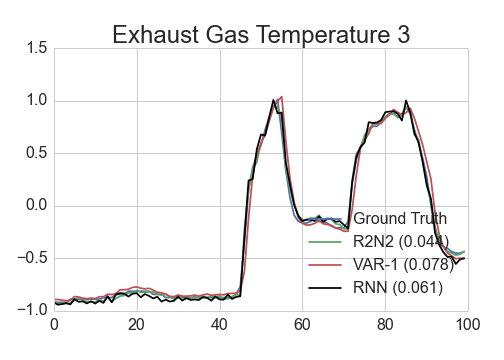}}
\subfloat[]{\includegraphics[width=0.45\linewidth]{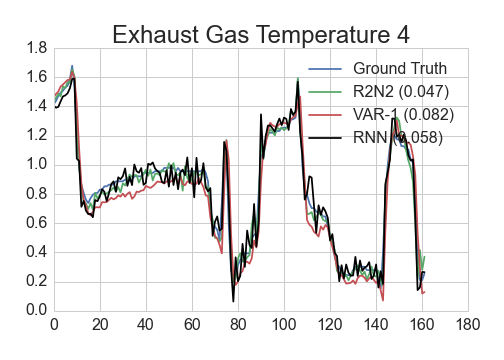}}
\vspace*{-8mm}
\caption{Plots of portions of features from a flight in the {\bf aviation data} and the 1-step predictions made by all models. Note that the x-axis and y-axis are different for each feature because selected sections of each feature have been zoomed-in for clarity. 8 features are shown here out of the total 42.}
\label{suppfig:aviation.pred.3}
\end{figure*}

\begin{figure*}
\vspace*{-12mm}
\centering
\captionsetup[subfigure]{font=small,aboveskip=0pt,belowskip=0pt,labelformat=empty}
\subfloat[]{\includegraphics[width=0.45\linewidth]{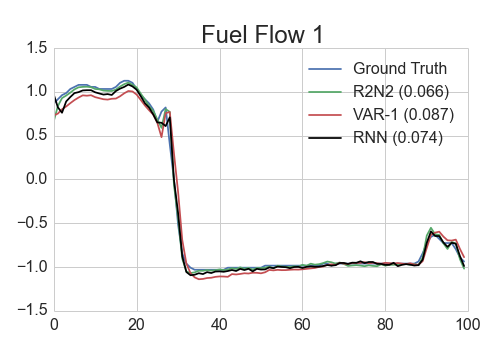}}
\subfloat[]{\includegraphics[width=0.45\linewidth]{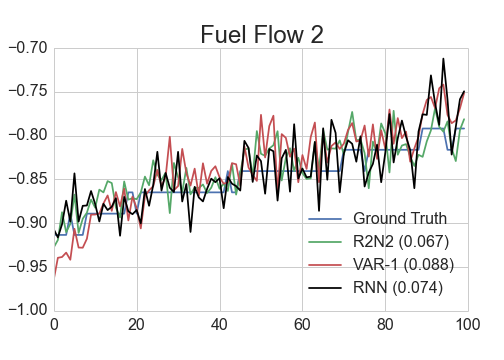}}
\vspace*{-12mm}
\\
\subfloat[]{\includegraphics[width=0.45\linewidth]{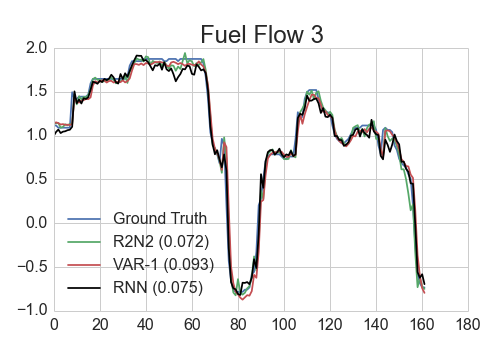}}
\subfloat[]{\includegraphics[width=0.45\linewidth]{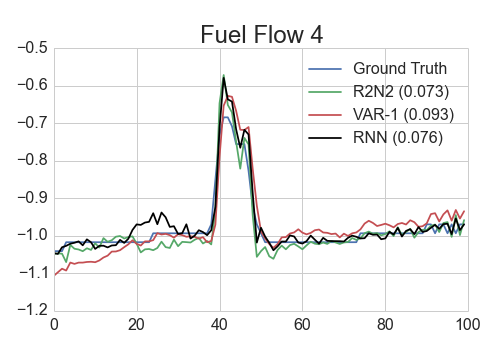}}
\vspace*{-12mm}
\\
\subfloat[]{\includegraphics[width=0.45\linewidth]{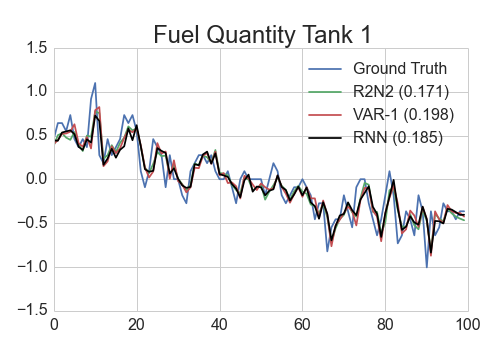}}
\subfloat[]{\includegraphics[width=0.45\linewidth]{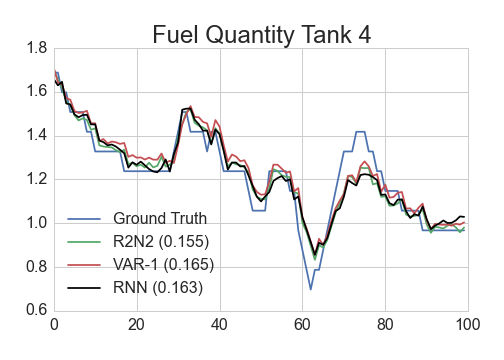}}
\vspace*{-12mm}
\\
\subfloat[]{\includegraphics[width=0.45\linewidth]{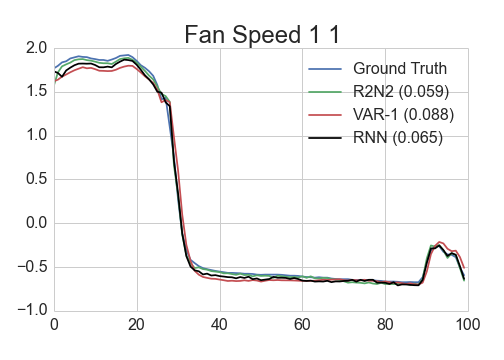}}
\subfloat[]{\includegraphics[width=0.45\linewidth]{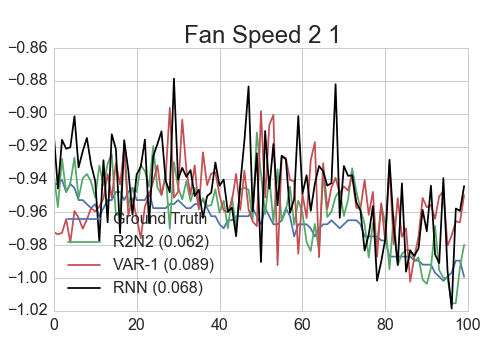}}
\vspace*{-8mm}

\caption{Plots of portions of features from a flight in the {\bf aviation data} and the 1-step predictions made by all models. Note that the x-axis and y-axis are different for each feature because selected sections of each feature have been zoomed-in for clarity. 16 features are shown here out of the total 42.}
\label{suppfig:aviation.pred.4}
\end{figure*}

\begin{figure*}
\vspace*{-12mm}
\centering
\captionsetup[subfigure]{font=small,aboveskip=0pt,belowskip=0pt,labelformat=empty}
\subfloat[]{\includegraphics[width=0.45\linewidth]{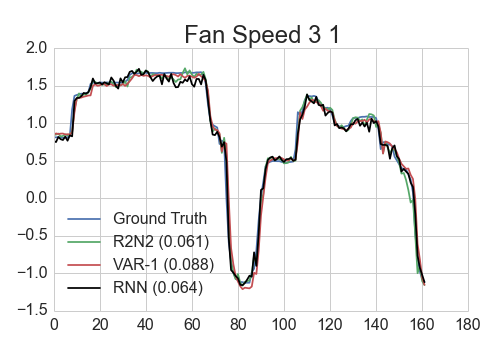}}
\subfloat[]{\includegraphics[width=0.45\linewidth]{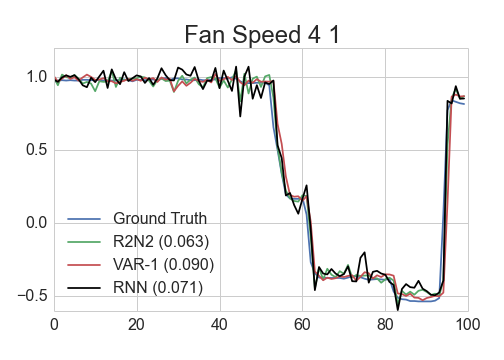}}
\vspace*{-12mm}
\\
\subfloat[]{\includegraphics[width=0.45\linewidth]{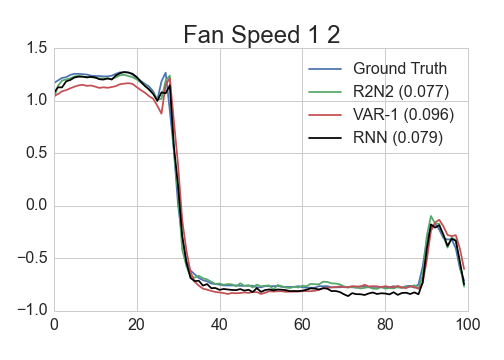}}
\subfloat[]{\includegraphics[width=0.45\linewidth]{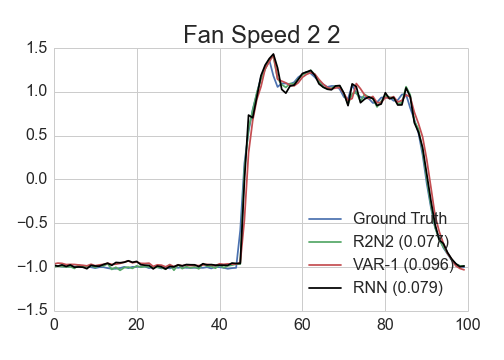}}
\vspace*{-12mm}
\\
\subfloat[]{\includegraphics[width=0.45\linewidth]{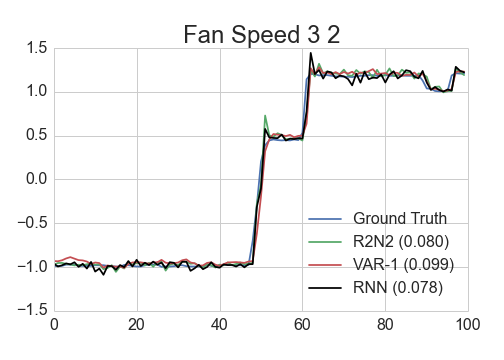}}
\subfloat[]{\includegraphics[width=0.45\linewidth]{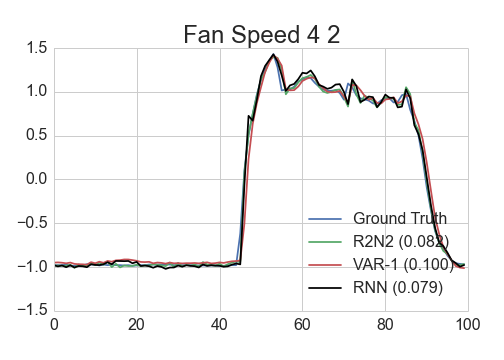}}
\vspace*{-12mm}
\\
\subfloat[]{\includegraphics[width=0.45\linewidth]{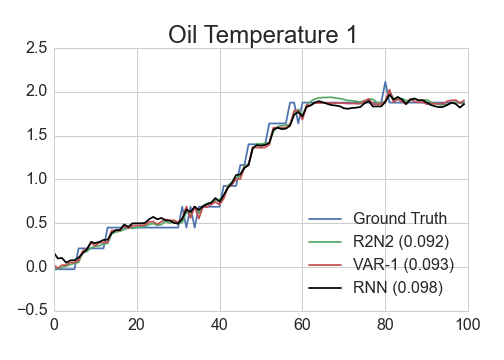}}
\subfloat[]{\includegraphics[width=0.45\linewidth]{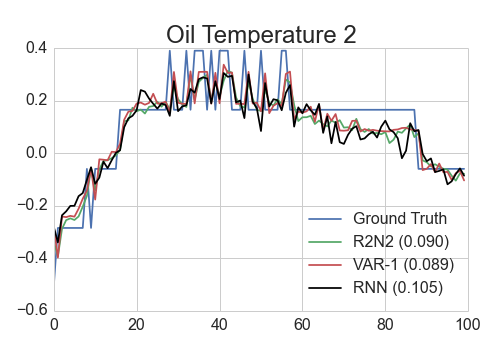}}
\vspace*{-8mm}

\caption{Plots of portions of features from a flight in the {\bf aviation data} and the 1-step predictions made by all models. Note that the x-axis and y-axis are different for each feature because selected sections of each feature have been zoomed-in for clarity. 8 features are shown here out of the total 42.}
\label{suppfig:aviation.pred.5}
\end{figure*}

\begin{figure*}
\vspace*{-12mm}
\centering
\captionsetup[subfigure]{font=small,aboveskip=0pt,belowskip=0pt,labelformat=empty}
\subfloat[]{\includegraphics[width=0.45\linewidth]{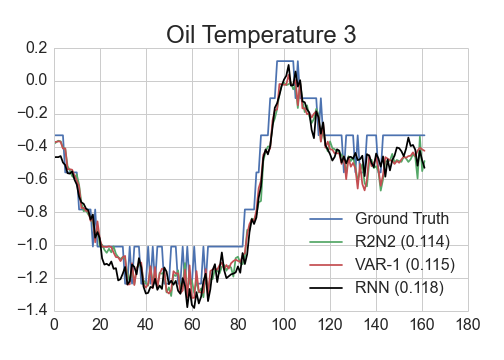}}
\subfloat[]{\includegraphics[width=0.45\linewidth]{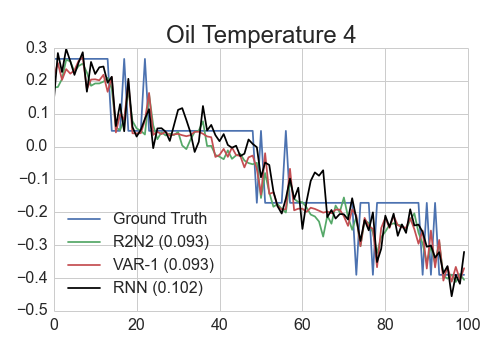}}
\vspace*{-8mm}
\caption{Plots of portions of features from a flight in the {\bf aviation data} and the 1-step predictions made by all models. Note that the x-axis and y-axis are different for each feature because selected sections of each feature have been zoomed-in for clarity. 8 features are shown here out of the total 42.}
\label{suppfig:aviation.pred.6}
\end{figure*}

Figures \ref{suppfig:aviation.pred.1}, \ref{suppfig:aviation.pred.2}, \ref{suppfig:aviation.pred.3}, \ref{suppfig:aviation.pred.4}, \ref{suppfig:aviation.pred.5} and \ref{suppfig:aviation.pred.6} show the 1-step ahead predictions by all models (VAR-1, RNN, R2N2) on a selected flight from the test set.

\section{Qualitative prediction plots of ENSO data}
\begin{figure*}
\vspace*{-12mm}
\centering
\captionsetup[subfigure]{font=small,aboveskip=0pt,belowskip=0pt,labelformat=empty}
\subfloat[]{\includegraphics[width=0.45\linewidth]{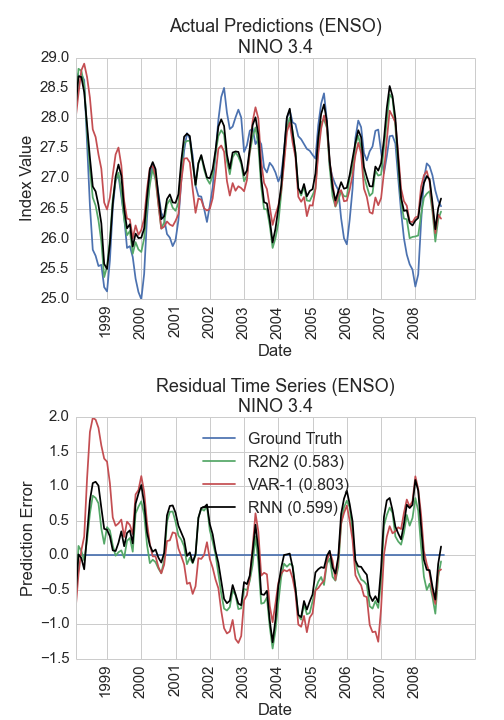}}
\subfloat[]{\includegraphics[width=0.45\linewidth]{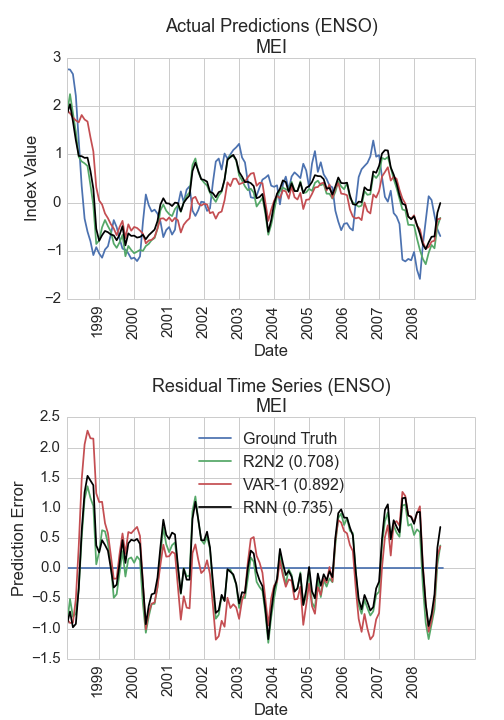}}
\vspace*{-12mm}
\\
\subfloat[]{\includegraphics[width=0.45\linewidth]{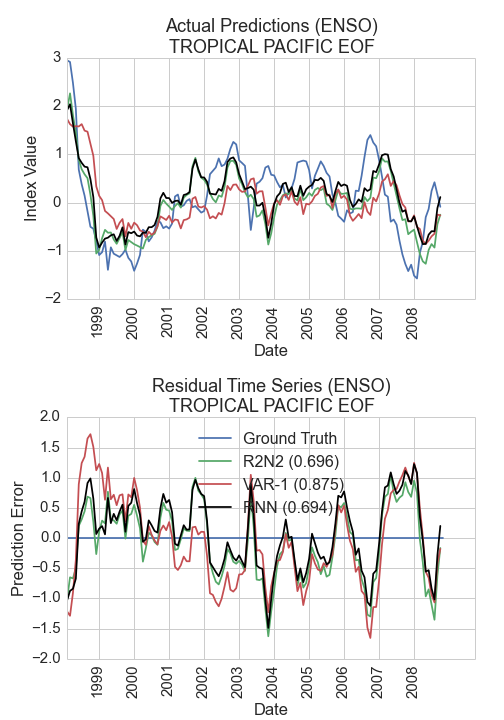}}
\subfloat[]{\includegraphics[width=0.45\linewidth]{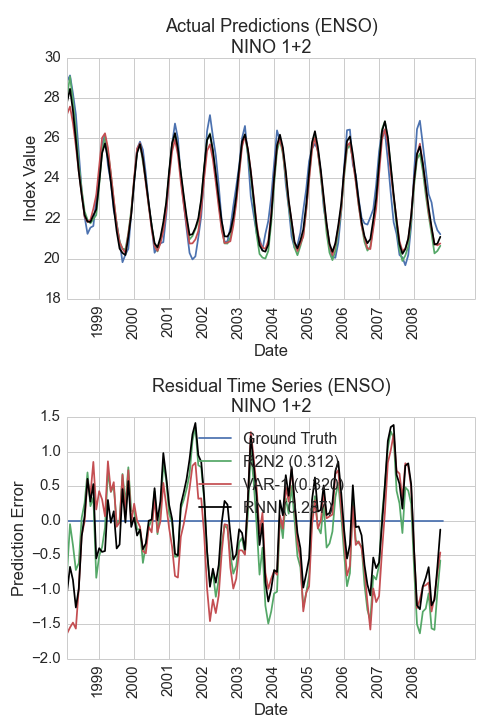}}
\vspace*{-8mm}

\caption{Plots of 4 out of 7 climate indices from the ENSO data, along with the predictions made by different models for $horizon = 6$. The top plot in each subfigure shows the actual predictions, while the bottom plot shows the prediction residuals.}
\label{suppfig:enso.pred.1}
\end{figure*}

\begin{figure*}
\vspace*{-12mm}
\centering
\captionsetup[subfigure]{font=small,aboveskip=0pt,belowskip=0pt,labelformat=empty}
\subfloat[]{\includegraphics[width=0.45\linewidth]{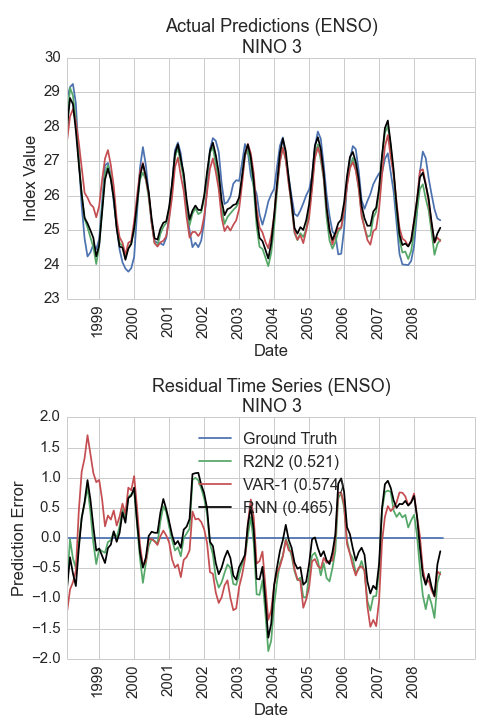}}
\subfloat[]{\includegraphics[width=0.45\linewidth]{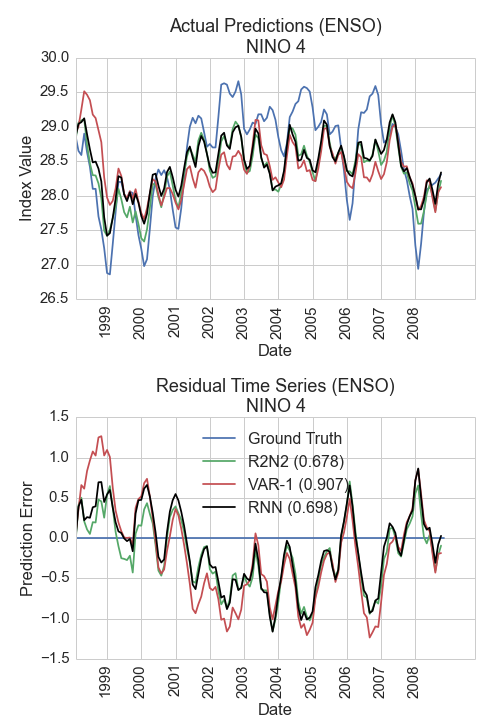}}
\vspace*{-12mm}
\\
\subfloat[]{\includegraphics[width=0.45\linewidth]{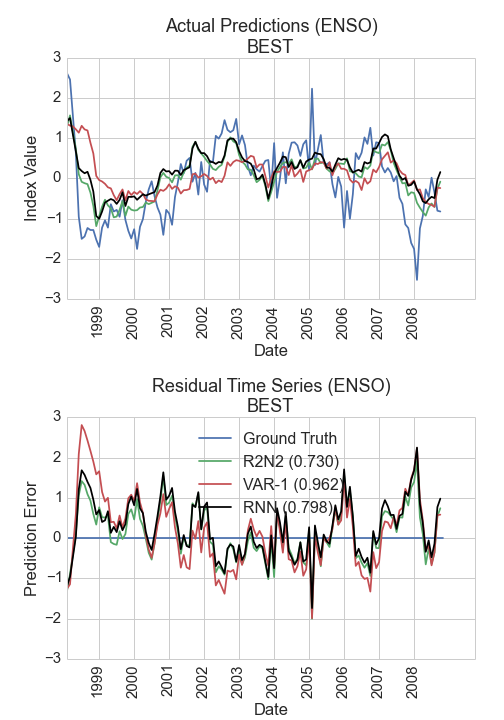}}
\vspace*{-8mm}

\caption{Plots of 3 out of 7 climate indices from the ENSO data, along with the predictions made by different models for $horizon = 6$. The top plot in each subfigure shows the actual predictions, while the bottom plot shows the prediction residuals.}
\label{suppfig:enso.pred.2}
\end{figure*}

Figures \ref{suppfig:enso.pred.1} and  \ref{suppfig:enso.pred.2} show the predictions on the ENSO data for $horizon = 6$, by all models (VAR-1, RNN, R2N2) along with the residual time-series plots.

\section{Aggregate performance plots}
\begin{figure*}[t]
\captionsetup[subfigure]{font=small,aboveskip=2pt,belowskip=-10pt,labelformat=empty}
\subfloat[]{\label{suppfig:aviation.mrse}\includegraphics[width=0.5\linewidth]{images/rse_aviation_var_rnn}}
\subfloat[]{\label{suppfig:aviation.re}\includegraphics[width=0.5\linewidth]{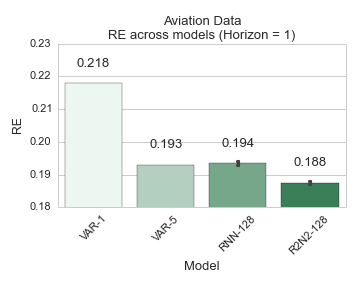}}

\caption{(a) The left figure shows the mean MRSE results on aviation test sets for all 3 models - VAR, RNN and R2N2. Note that for VAR we show both order-1 results and the best order results (which was 5 in this case). RNN-128 means an RNN with a hidden layer size of 128. R2N2-128 uses VAR-1 as the base model and an RNN with hidden layer size of 128. R2N2-128 performs the best among all the model architectures. (b) The figure on the right shows the comparison of RE values for the models. Since the aviation data has zero mean, the values of MRSE and RE are the same. The error bars in both the figures represent the variability due to different initializations of parameters across multiple runs. The bars are small indicating good stability in performance for both RNN and R2N2.}
\label{suppfig:aviation.rse}
\end{figure*}

\begin{figure*}
\captionsetup[subfigure]{font=small,aboveskip=2pt,belowskip=-10pt,labelformat=empty}
\subfloat[]{\label{suppfig:enso.mrse}\includegraphics[width=0.5\linewidth]{images/enso_main/enso_rse_models_1}}
\subfloat[]{\label{suppfig:enso.re}\includegraphics[width=0.5\linewidth]{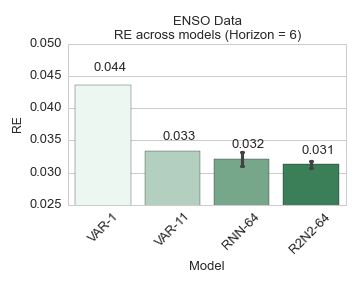}}

\caption{(a) The left figure shows the MRSE results on the ENSO data for all 3 models - VAR, RNN and R2N2. Note that for VAR we show both order-1 results and the best order results (which was 11 in this case). RNN-64 means an RNN with a hidden layer size of 64. R2N2-64 uses VAR-1 as the base model and an RNN with hidden layer size of 64. R2N2-64 performs the best among all the model architectures. (b) The figure on the right shows the comparison of RE values for the models. The scale of the values is very small because the original data is on a much larger scale. Nevertheless, a small gain is still seen in the R2N2-64 performance.}
\label{suppfig:enso.rse}

\end{figure*}

Figure \ref{suppfig:aviation.rse} shows the aggregate quantitative performance of both the metrics MRSE and RE on the aviation dataset. Figure \ref{suppfig:enso.rse} shows the same for the ENSO dataset. Note that both the metrics show similar relative behavior among models. RE is same as MRSE for the aviation dataset because the mean feature values for this data is 0.

\section{Architecture comparison : RNN vs R2N2}

\begin{figure*}
\captionsetup[subfigure]{font=small,aboveskip=2pt,belowskip=-10pt}
\subfloat[]{\label{suppfig:aviation.rse.hidden}\includegraphics[width=0.5\linewidth]{images/rse_rnn_vs_r2n2_hidden}}
\subfloat[]{\label{suppfig:aviation.corr.hidden}\includegraphics[width=0.5\linewidth]{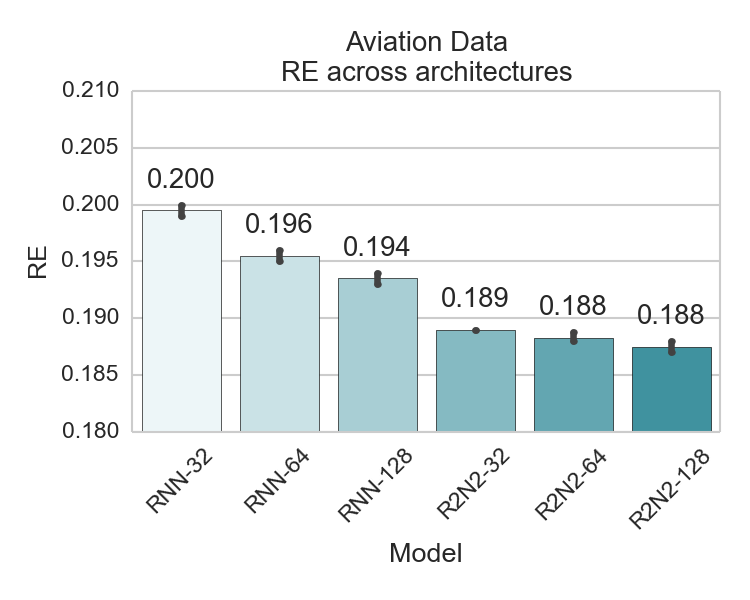}}

\vspace{1mm}

\caption{These figures show the comparison of MRSE and RE values across different hidden layer sizes of RNN and R2N2 for the aviation data. Note that even R2N2-32 performs better than RNN-128. This implies that very low complexity RNNs can be used as components in R2N2, which provides savings in terms of computational complexity.}
\label{suppfig:aviation.hidden}
\end{figure*}

Figure \ref{suppfig:aviation.hidden} shows the behavior of MRSE and RE, on aviation test sets, as the sizes of the hidden layer is changed in RNN and R2N2. Again, the values of RE are the same as MRSE because the feature means are 0 for the aviation dataset.

\end{document}